# Generalized Principal Component Analysis (GPCA)

René Vidal, *Member, IEEE*, Yi Ma, *Member, IEEE*, and Shankar Sastry, *Fellow, IEEE*

**Abstract**—This paper presents an algebro-geometric solution to the problem of segmenting an unknown number of subspaces of unknown and varying dimensions from sample data points. We represent the subspaces with a set of homogeneous polynomials whose degree is the number of subspaces and whose derivatives at a data point give normal vectors to the subspace passing through the point. When the number of subspaces is known, we show that these polynomials can be estimated linearly from data; hence, subspace segmentation is reduced to classifying one point per subspace. We select these points optimally from the data set by minimizing certain distance function, thus dealing automatically with moderate noise in the data. A basis for the complement of each subspace is then recovered by applying standard PCA to the collection of derivatives (normal vectors). Extensions of GPCA that deal with data in a high-dimensional space and with an unknown number of subspaces are also presented. Our experiments on low-dimensional data show that GPCA outperforms existing algebraic algorithms based on polynomial factorization and provides a good initialization to iterative techniques such as K-subspaces and Expectation Maximization. We also present applications of GPCA to computer vision problems such as face clustering, temporal video segmentation, and 3D motion segmentation from point correspondences in multiple affine views.

## 1 Introduction

Principal Component Analysis (PCA) [12] refers to the problem of fitting a linear subspace $S \subset \mathbb{R}^D$ of unknown dimension $d < D$ to $N$ sample points $\{\boldsymbol{x}_j\}_{j=1}^N$ in $S$. This problem shows up in a variety of applications in many fields, e.g., pattern recognition, data compression, regression, image processing, etc., and can be solved in a remarkably simple way from the singular value decomposition (SVD) of the (mean-subtracted) data matrix $[\boldsymbol{x}_1, \boldsymbol{x}_2, \dots, \boldsymbol{x}_N] \in \mathbb{R}^{D\times N}$.[1] With noisy data, this linear algebraic solution has the geometric interpretation of minimizing the sum of the squared distances from the (noisy) data points $\boldsymbol{x}_j$ to their projections $\tilde{\boldsymbol{x}}^j$ in $S$.

In addition to these algebraic and geometric interpretations, PCA can also be understood in a probabilistic manner. In Probabilistic PCA [20] (PPCA), the noise is assumed to be drawn from an unknown distribution and the problem becomes one of identifying the subspace and distribution parameters in a maximum-likelihood sense. When the noise distribution is Gaussian, the algebro-geometric and probabilistic interpretations coincide [2]. However, when the noise distribution is non-Gaussian, the solution to PPCA is no longer linear, as shown in [2], where PCA is generalized to arbitrary distributions in the exponential family.

Another extension of PCA is nonlinear principal components (NLPCA) or Kernel PCA (KPCA), which is the problem of identifying a *nonlinear* manifold from sample points. The standard solution to NLPCA [16] is based on first embedding the data into a higher-dimensional *feature* space $\mathbf{F}$ and then applying standard PCA to the embedded data. Since the dimension of $\mathbf{F}$ can be large, a more practical solution is obtained from the eigenvalue decomposition of the so-called *kernel* matrix; hence, the name KPCA. One of the disadvantages of KPCA is that, in practice, it is difficult to determine which kernel function to use because the choice of the kernel naturally depends on the nonlinear structure of the manifold to be identified. In fact, learning kernels is an active topic of research in machine learning. To the best of our knowledge, our work is the first one to prove analytically that the Veronese map (a polynomial embedding) is the natural embedding for data lying in a union of multiple subspaces.

In this paper, we consider the following alternative extension of PCA to the case of data lying in a union of subspaces, as illustrated in Fig. 1 for two subspaces of $\mathbb{R}^3$.

**Problem (Subspace Segmentation).** *Given a set of points* $\boldsymbol{X} = \{\boldsymbol{x}_j \in \mathbb{R}^D\}_{j=1}^N$ *drawn from* $n \geq 1$ *different linear subspaces* $\{S_i \subseteq \mathbb{R}^D\}_{i=1}^n$ *of dimension* $d_i = \dim(S_i)$, $0 < d_i < D$, *without knowing which points belong to which subspace:*

1. *find the number of subspaces* $n$ *and their dimensions* $\{d_i\}_{i=1}^n$,
2. *find a basis for each subspace* $S_i$ *(or for* $S_i^\perp$*), and*
3. *group the* $N$ *sample points into the* $n$ *subspaces.*

1. In the context of stochastic signal processing, PCA is also known as the Karhunen-Loeve transform [18]; in the applied statistics literature, SVD is also known as the Eckart and Young decomposition [4].

- *R. Vidal is with the Center for Imaging Science, Department of Biomedical Engineering, The Johns Hopkins University, 308B Clark Hall, 3400 N. Charles Street, Baltimore, MD 21218. E-mail: rvidal@cis.jhu.edu.*
- *Y. Ma is with the Electrical and Computer Engineering Department, University of Illinois at Urbana-Champaign, 145 Coordinated Science Laboratory, 1308 West Main Street, Urbana, IL 61801. E-mail: yima@uiuc.edu.*
- *S. Sastry is with the Department of Electrical Engineering and Computer Sciences, University of California, Berkeley, 514 Cory Hall, Berkeley, CA 94720. E-mail: sastry@eecs.berkeley.edu.*

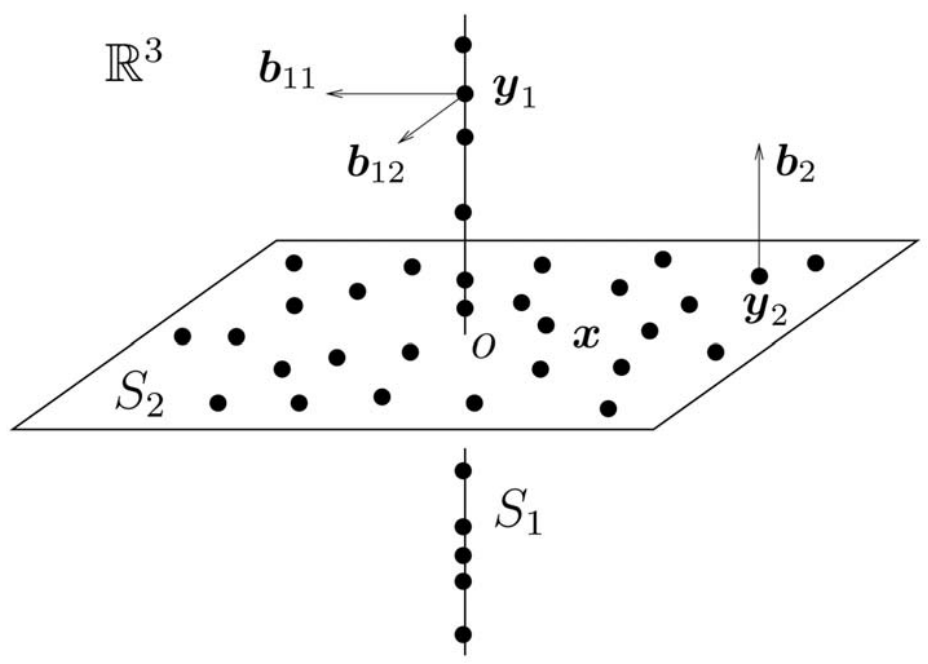

Fig. 1. Data points drawn from the union of one plane and one line (through the origin $o$) in $\mathbb{R}^3$. The objective of subspace segmentation is to identify the normal vectors $b_{11}$, $b_{12}$, and $b_2$ to each one of the subspaces from the data.

### 1.1 Previous Work on Subspace Segmentation

Subspace segmentation is a fundamental problem in many applications in computer vision (e.g., image/motion/video segmentation), image processing (e.g., image representation and compression), and systems theory (e.g., hybrid system identification), which is usually regarded as "chicken-and-egg." If the segmentation of the data was known, one could easily fit a single subspace to each group of points using standard PCA. Conversely, if the subspace bases were known, one could easily find the data points that best fit each subspace. Since, in practice, neither the subspace bases nor the segmentation of the data are known, most existing methods randomly choose a basis for each subspace and then iterate between data segmentation and subspace estimation. This can be done using, e.g., K-subspaces [10], an extension of K-means to the case of subspaces, subspace growing and subspace selection [15], or Expectation Maximization (EM) for mixtures of PCAs [19]. Unfortunately, most iterative methods are, in general, very sensitive to initialization; hence, they may not converge to the global optimum [21].

The need for initialization methods has motivated the recent development of algebro-geometric approaches to subspace segmentation that do *not* require initialization. In [13] (see, also, [3]), it is shown that when the subspaces are orthogonal, of equal dimension $d$, and intersect only at the origin, which implies that $D \geq nd$, one can use the SVD of the data to define a similarity matrix from which the segmentation of the data can be obtained using spectral clustering techniques. Unfortunately, this method is sensitive to noise in the data, as shown in [13], [27] where various improvements are proposed, and fails when the subspaces intersect arbitrarily [14], [22], [28]. The latter case has been addressed in an ad hoc fashion by using clustering algorithms such as K-means, spectral clustering, or EM [14], [28] to segment the data and PCA to obtain a basis for each group. The only algebraic approaches that deal with arbitrary intersections are [17], which studies the case of two planes in $\mathbb{R}^3$ and [24] which studies the case of subspaces of codimension one, i.e., *hyperplanes*, and shows that hyperplane segmentation is equivalent to homogeneous polynomial factorization. Our previous work [23] extended this framework to subspaces of unknown and possibly different dimensions under the additional assumption that the number of subspaces is known. This paper unifies the results of [24] and [23] and extends to the case in which both the number and dimensions of the subspaces are unknown.

### 1.2 Paper Organization and Contributions

In this paper, we propose an algebro-geometric approach to subspace segmentation called *Generalized Principal Component Analysis* (GPCA), which is based on fitting, differentiating, and dividing polynomials. Unlike prior work, we do not restrict the subspaces to be orthogonal, trivially intersecting, or with known and equal dimensions. Instead, we address the most general case of an *arbitrary number of subspaces* of *unknown* and possibly *different* dimensions (e.g., Fig. 1) and with *arbitrary intersections* among the subspaces.

In Section 2, we motivate and highlight the key ideas of our approach by solving the simple example shown in Fig. 1.

In Section 3, we generalize this example to the case of data lying in a known number of subspaces with unknown and possibly different dimensions. We show that one can represent the union of all subspaces as the zero set of a collection of homogeneous polynomials whose degree is the number of subspaces and whose factors encode normal vectors to the subspaces. The coefficients of these polynomials can be linearly estimated from sample data points on the subspaces and the set of normal vectors to each subspace can be obtained by evaluating the derivatives of these polynomials at any point lying on the subspace. Therefore, subspace segmentation is reduced to the problem of classifying one point per subspace. When those points are given (e.g., in semisupervised learning), this means that in order to learn the mixture of subspaces, it is sufficient to have *one positive example per class*. When all the data points are unlabeled (e.g., in unsupervised learning), we use polynomial division to recursively select points in the data set that minimize their distance to the algebraic set; hence, dealing automatically with moderate noise in the data. A basis for the complement of each subspace is then recovered by applying standard PCA to the derivatives of the polynomials (normal vectors) at those points. The final result is a global, noniterative subspace segmentation algorithm based on simple linear and polynomial algebra.

In Section 4, we discuss some extensions of our approach. We show how to deal with low-dimensional subspaces of a high-dimensional space via a linear projection onto a low-dimensional subspace that preserves the number and dimensions of the subspaces. We also show how to generalize the basic GPCA algorithm to the case in which the number of subspaces is unknown via a recursive partitioning algorithm.

In Section 5, we present experiments on low-dimensional data showing that GPCA gives about half the error of existing algebraic algorithms based on polynomial factorization, and improves the performance of iterative techniques, such as K-subspaces and EM, by about 50 percent with respect to random initialization. We also present applications of GPCA to computer vision problems such as face clustering, temporal video segmentation, and 3D motion segmentation from point correspondences in multiple affine views.

## 2 An Introductory Example

Imagine that we are given data in $\mathbb{R}^3$ drawn from a line $S_1 = \{\boldsymbol{x} : x_1 = x_2 = 0\}$ and a plane $S_2 = \{\boldsymbol{x} : x_3 = 0\}$, as shown in Fig. 1. We can describe the two subspaces as

$$\begin{aligned} S_1 \cup S_2 &= \{\boldsymbol{x} : (x_1 = x_2 = 0) \vee (x_3 = 0)\} \\ &= \{\boldsymbol{x} : (x_1 x_3 = 0) \wedge (x_2 x_3 = 0)\}. \end{aligned}$$

Therefore, even though each individual subspace is described with polynomials of degree one (linear equations), the mixture of two subspaces is described with two polynomials of degree two, namely, $p_{21}(\boldsymbol{x}) = x_1x_3$ and $p_{22}(\boldsymbol{x}) = x_2x_3$. More generally, any two linear subspaces in $\mathbb{R}^3$ can be represented as the set of points satisfying some polynomials of the form

$$c_1x_1^2 + c_2x_1x_2 + c_3x_1x_3 + c_4x_2^2 + c_5x_2x_3 + c_6x_3^2 = 0.$$

Although these polynomials are nonlinear in each data point $[x_1, x_2, x_3]^T$, they are actually linear in the coefficient vector $\boldsymbol{c} = [c_1, \ldots, c_6]^T$. Therefore, given enough data points, one can linearly *fit* these *polynomials* to the *data*.

Given the collection of polynomials that vanish on the data points, we would like to compute a basis for each subspace. In our example, let $P_2(\boldsymbol{x}) = [p_{21}(\boldsymbol{x}), p_{22}(\boldsymbol{x})]$ and consider the derivatives of $P_2(\boldsymbol{x})$ at two points in each of the subspaces $\boldsymbol{y}_1 = [0,0,1]^T \in S_1$ and $\boldsymbol{y}_2 = [1,1,0]^T \in S_2$:

$$DP_2(\boldsymbol{x}) = \begin{bmatrix} x_3 & 0 \\ 0 & x_3 \\ x_1 & x_2 \end{bmatrix} \Rightarrow$$

$$DP_2(\boldsymbol{y}_1) = \begin{bmatrix} 1 & 0 \\ 0 & 1 \\ 0 & 0 \end{bmatrix}, DP_2(\boldsymbol{y}_2) = \begin{bmatrix} 0 & 0 \\ 0 & 0 \\ 1 & 1 \end{bmatrix}.$$

Note that the columns of $DP_2(\boldsymbol{y}_1)$ span $S_1^\perp$ and the columns of $DP_2(\boldsymbol{y}_2)$ span $S_2^\perp$ (see Fig. 1). Also, the dimension of the line is $d_1 = 3 - \text{rank}(DP_2(\boldsymbol{y}_1)) = 1$ and the dimension of the plane is $d_2 = 3 - \text{rank}(DP_2(\boldsymbol{y}_2)) = 2$. Thus, if we are given one point in each subspace, we can obtain the *subspace bases* and their *dimensions* from the *derivatives of the polynomials* at these points.

The final question is to find one point per subspace, so that we can obtain the normal vectors from the derivatives of $P_2$ at those points. With perfect data, we may choose a first point as any of the points in the data set. With noisy data, we may first define a distance from any point in $\mathbb{R}^3$ to one of the subspaces, e.g., the algebraic distance $d_2(\boldsymbol{x})^2 = p_{21}(\boldsymbol{x})^2 + p_{22}(\boldsymbol{x})^2 = (x_1^2 + x_2^2)x_3^2$, and then choose a point in the data set that minimizes this distance. Say, we pick $\boldsymbol{y}_2 \in S_2$ as such point. We can then compute the normal vector $\boldsymbol{b}_2 = [0,0,1]^T$ to $S_2$ from $DP(\boldsymbol{y}_2)$. As it turns out, we can pick a second point in $S_1$ but not in $S_2$ by *polynomial division*. We can just divide the original polynomials of degree $n = 2$ by $(\boldsymbol{b}_2^T\boldsymbol{x})$ to obtain polynomials of degree $n - 1 = 1$:

$$p_{11}(\boldsymbol{x}) = \frac{p_{21}(\boldsymbol{x})}{\boldsymbol{b}_2^T\boldsymbol{x}} = x_1 \quad \text{and} \quad p_{12}(\boldsymbol{x}) = \frac{p_{22}(\boldsymbol{x})}{\boldsymbol{b}_2^T\boldsymbol{x}} = x_2.$$

Since these new polynomials vanish on $S_1$ but not on $S_2$, we can find a point $\boldsymbol{y}_1$ in $S_1$ but not in $S_2$, as a point in the data set that minimizes $d_1(\boldsymbol{x})^2 = p_{11}(\boldsymbol{x})^2 + p_{12}(\boldsymbol{x})^2 = x_1^2 + x_2^2$.

As we will show in the next section, one can also solve the more general problem of segmenting a union of $n$ subspaces $\{S_i \subset \mathbb{R}^D\}_{i=1}^n$ of *unknown* and possibly *different* dimensions $\{d_i\}_{i=1}^n$ by *polynomial fitting* (Section 3.3), *differentiation* (Section 3.4), and *division* (Section 3.5).

# 3 Generalized Principal Component Analysis

In this section, we derive a constructive algebro-geometric solution to the subspace segmentation problem when the number of subspaces $n$ is *known*. The case in which the number of subspaces is unknown will be discussed in Section 4.2. Our algebro-geometric solution is summarized in the following theorem:

**Theorem 1 (Generalized Principal Component Analysis).** *A union of $n$ subspaces of $\mathbb{R}^D$ can be represented with a set of homogeneous polynomials of degree $n$ in $D$ variables. These polynomials can be estimated linearly given enough sample points in general position in the subspaces. A basis for the complement of each subspace can be obtained from the derivatives of these polynomials at a point in each of the subspaces. Such points can be recursively selected via polynomial division. Therefore, the subspace segmentation problem is mathematically equivalent to fitting, differentiating and dividing a set of homogeneous polynomials.*

## 3.1 Notation

Let $\boldsymbol{x}$ be a vector in $\mathbb{R}^D$. A homogeneous polynomial of degree $n$ in $\boldsymbol{x}$ is a polynomial $p_n(\boldsymbol{x})$ such that $p_n(\lambda\boldsymbol{x}) = \lambda^n p_n(\boldsymbol{x})$ for all $\lambda$ in $\mathbb{R}$. The space of all homogeneous polynomials of degree $n$ in $D$ variables is a vector space of dimension $M_n(D) = \binom{n+D-1}{D-1}$. A particular basis for this space is given by all the monomials of degree $n$ in $D$ variables, that is $\boldsymbol{x}^I = x_1^{n_1}x_2^{n_2}\cdots x_D^{n_D}$ with $0 \le n_j \le n$ for $j = 1, \ldots, D$, and $n_1 + n_2 + \cdots + n_D = n$. Thus, each homogeneous polynomial can be written as a linear combination of the monomials $\boldsymbol{x}^I$ with coefficient vector $\boldsymbol{c}_n \in \mathbb{R}^{M_n(D)}$ as

$$p_n(\boldsymbol{x}) = \boldsymbol{c}_n^T\nu_n(\boldsymbol{x}) = \sum c_{n_1,n_2,\ldots,n_D}x_1^{n_1}x_2^{n_2}\cdots x_D^{n_D}, \tag{1}$$

where $\nu_n : \mathbb{R}^D \rightarrow \mathbb{R}^{M_n(D)}$ is the *Veronese map* of degree $n$ [7], also known as the *polynomial embedding* in machine learning, defined as $\nu_n : [x_1, \ldots, x_D]^T \mapsto [\ldots, \boldsymbol{x}^I, \ldots]^T$ with $I$ chosen in the degree-lexicographic order.

**Example 1 (The Veronese map of degree 2 in three variables).** If $\boldsymbol{x} = [x_1, x_2, x_3]^T \in \mathbb{R}^3$, the Veronese map of degree 2 is given by:

$$\nu_2(\boldsymbol{x}) = [x_1^2, x_1x_2, x_1x_3, x_2^2, x_2x_3, x_3^2]^T \in \mathbb{R}^6. \tag{2}$$

## 3.2 Representing a Union of $n$ Subspaces with a Set of Homogeneous Polynomials of Degree $n$

We represent a subspace $S_i \subset \mathbb{R}^D$ of dimension $d_i$, where $0 < d_i < D$, by choosing a basis

$$B_i \doteq [\boldsymbol{b}_{i1}, \ldots, \boldsymbol{b}_{i(D-d_i)}] \in \mathbb{R}^{D\times(D-d_i)} \tag{3}$$

for its orthogonal complement $S_i^\perp$. One could also choose a basis for $S_i$ directly, especially if $d_i \ll D$. Section 4.1 will show that the problem can be reduced to the case $D = \max\{d_i\} + 1$; hence, the orthogonal representation is more convenient if $\max\{d_i\}$ is small. With this representation, each subspace can be expressed as the set of points satisfying $D - d_i$ linear equations (polynomials of degree *one*), that is,

$$S_i = \{\boldsymbol{x} \in \mathbb{R}^D : B_i^T\boldsymbol{x} = 0\} = \Big\{\boldsymbol{x} \in \mathbb{R}^D : \bigwedge_{j=1}^{D-d_i} (\boldsymbol{b}_{ij}^T\boldsymbol{x} = 0)\Big\}. \tag{4}$$

For affine subspaces (which do not necessarily pass through the origin), we use homogeneous coordinates so that they become linear subspaces.

We now demonstrate that one can represent the union of $n$ subspaces $\{S_i \subset \mathbb{R}^D\}_{i=1}^n$ with a set of polynomials whose degree is $n$ rather than one. To see this, notice that $\boldsymbol{x} \in \mathbb{R}^D$ belongs to $\cup_{i=1}^n S_i$ if and only if it satisfies $(\boldsymbol{x} \in S_1) \vee \ldots \vee (\boldsymbol{x} \in S_n)$. This is equivalent to

$$\bigvee_{i=1}^{n} (\boldsymbol{x} \in S_i) \Leftrightarrow \bigvee_{i=1}^{n} \bigwedge_{j=1}^{D-d_i} (\boldsymbol{b}_{ij}^T \boldsymbol{x} = 0) \Leftrightarrow \bigwedge_{\sigma} \bigvee_{i=1}^{n} (\boldsymbol{b}_{i\sigma(i)}^T \boldsymbol{x} = 0), \quad (5)$$

where the right-hand side is obtained by exchanging ands and ors using De Morgan's laws and $\sigma$ is a particular choice of one normal vector $\boldsymbol{b}_{i\sigma(i)}$ from each basis $B_i$. Since each one of the $\prod_{i=1}^n (D - d_i)$ equations in (5) is of the form

$$\bigvee_{i=1}^{n} (\boldsymbol{b}_{i\,\sigma(i)}^T \boldsymbol{x} = 0) \Leftrightarrow \Big( \prod_{i=1}^{n} (\boldsymbol{b}_{i\,\sigma(i)}^T \boldsymbol{x}) = 0 \Big) \Leftrightarrow (p_{n\sigma}(\boldsymbol{x}) = 0), \quad (6)$$

i.e., a homogeneous polynomial of degree $n$ in $D$ variables, we can write each polynomial as a linear combination of monomials $\boldsymbol{x}^I$ with coefficient vector $\boldsymbol{c}_n \in \mathbb{R}^{M_n(D)}$, as in (1). Therefore, we have the following result.

**Theorem 2 (Representing Subspaces with Polynomials).** *A union of $n$ subspaces can be represented as the set of points satisfying a set of homogeneous polynomials of the form*

$$p_n(\boldsymbol{x}) = \prod_{i=1}^{n} (\boldsymbol{b}_i^T \boldsymbol{x}) = \boldsymbol{c}_n^T \nu_n(\boldsymbol{x}) = 0, \quad (7)$$

*where $\boldsymbol{b}_i \in \mathbb{R}^D$ is a normal vector to the $i$th subspace.*

The importance of Theorem 2 is that it allows us to solve the "chicken-and-egg" problem described in Section 1.1 algebraically, because the polynomials in (7) are satisfied by *all* data points, regardless of which point belongs to which subspace. We can then use all the data to estimate all the subspaces, without prior segmentation and without having to iterate between data segmentation and model estimation, as we will show in Sections 3.3, 3.4, and 3.5.

### 3.3 Fitting Polynomials to Data Lying in Multiple Subspaces

Thanks to Theorem 2, the problem of identifying a union of $n$ subspaces $\{S_i\}_{i=1}^n$ from a set of data points $\boldsymbol{X} \doteq \{\boldsymbol{x}_j\}_{j=1}^N$ lying in the subspaces is equivalent to solving for the normal bases $\{B_i\}_{1=1}^n$ from the set of *nonlinear* equations in (6). Although these polynomial equations are nonlinear in each data point $\boldsymbol{x}$, they are actually *linear* in the coefficient vector $\boldsymbol{c}_n$. Indeed, since each polynomial $p_n(\boldsymbol{x}) = \boldsymbol{c}_n^T \nu_n(\boldsymbol{x})$ must be satisfied by every data point, we have $\boldsymbol{c}_n^T \nu_n(\boldsymbol{x}_j) = 0$ for all $j = 1, \ldots, N$. We use $I_n$ to denote the space of coefficient vectors $\boldsymbol{c}_n$ of all homogeneous polynomials that vanish on the $n$ subspaces. Obviously, the coefficient vectors of the factorizable polynomials defined in (6) span a (possibly proper) subspace in $I_n$:

$$\mathrm{span}_\sigma\{p_{n\sigma}\} \subseteq I_n. \quad (8)$$

As every vector $\boldsymbol{c}_n$ in $I_n$ represents a polynomial that vanishes on all the data points (on the subspaces), the vector must satisfy the system of linear equations

$$\boldsymbol{c}_n^T \boldsymbol{V}_n(D) \doteq \boldsymbol{c}_n^T [\, \nu_n(\boldsymbol{x}_1) \quad \ldots \quad \nu_n(\boldsymbol{x}_N) \,] = \boldsymbol{0}^T. \quad (9)$$

$\boldsymbol{V}_n(D) \in \mathbb{R}^{M_n(D) \times N}$ is called the *embedded data matrix*. Obviously, we have the relationship

$$I_n \subseteq \mathrm{null}(\boldsymbol{V}_n(D)).$$

Although we know that the coefficient vectors $\boldsymbol{c}_n$ of vanishing polynomials must lie in the left null space of $\boldsymbol{V}_n(D)$, we do not know if every vector in the null space corresponds to a polynomial that vanishes on the subspaces. Therefore, we would like to study under what conditions on the data points, we can solve for the unique $m_n \doteq \dim(I_n)$ independent polynomials that vanish on the subspaces from the null space of $\boldsymbol{V}_n$. Clearly, a necessary condition is to have $N \geq \sum_{i=1}^n d_i$ points in $\cup_{i=1}^n S_i$, with at least $d_i$ points in general position within each subspace $S_i$, i.e., the $d_i$ points must span $S_i$. However, because we are representing each polynomial $p_n(\boldsymbol{x})$ linearly via the coefficient vector $\boldsymbol{c}_n$, we need a number of samples such that a basis for $I_n$ can be uniquely recovered from $\mathrm{null}(\boldsymbol{V}_n(D))$. That is, the number of samples $N$ must be such that

$$\mathrm{rank}(\boldsymbol{V}_n(D)) = M_n(D) - m_n \leq M_n(D) - 1. \quad (10)$$

Therefore, if the number of subspaces $n$ is known, we can recover $I_n$ from $\mathrm{null}(\boldsymbol{V}_n(D))$ given $N \geq M_n(D) - 1$ points in general position. A basis of $I_n$ can be computed *linearly* as the set of $m_n$ left singular vectors of $\boldsymbol{V}_n(D)$ associated with its $m_n$ zero singular values. Thus, we obtain a basis of polynomials of degree $n$, say $\{p_{n\ell}\}_{\ell=1}^{m_n}$, that vanish on the $n$ subspaces.

**Remark 1 (GPCA and Kernel PCA).** Kernel PCA identifies a manifold from sample data by embedding the data into a higher-dimensional feature space $\boldsymbol{F}$ such that the embedded data points lie in a linear subspace of $\boldsymbol{F}$. Unfortunately, there is no general methodology for finding the appropriate embedding for a particular problem because the embedding naturally depends on the geometry of the manifold. The above derivation shows that the commonly used *polynomial embedding* $\nu_n$ is the appropriate embedding to use in KPCA when the original data lie in a union of subspaces, because the embedded data points $\{\nu_n(\boldsymbol{x}_j)\}_{j=1}^N$ lie in a subspace of $\mathbb{R}^{M_n(D)}$ of dimension $M_n(D) - m_n$, where $m_n = \dim(I_n)$. Notice also that the matrix $\mathcal{C} = \boldsymbol{V}_n(D)\boldsymbol{V}_n(D)^T \in \mathbb{R}^{M_n(D) \times M_n(D)}$ is exactly the *covariance matrix* in the feature space and $\mathcal{K} = \boldsymbol{V}_n(D)^T \boldsymbol{V}_n(D) \in \mathbb{R}^{N \times N}$ is the *kernel matrix* associated with the $N$ embedded samples.

**Remark 2 (Estimation from Noisy Data).** In the presence of moderate noise, we can still estimate the coefficients of each polynomial in a least-squares sense as the singular vectors of $\boldsymbol{V}_n(D)$ associated with its smallest singular values. However, we cannot directly estimate the number of polynomials from the rank of $\boldsymbol{V}_n(D)$ because $\boldsymbol{V}_n(D)$ may be of full rank. We use model selection to determine $m_n$ as

$$m_n = \arg\min_m \frac{\sigma_{m+1}^2(\boldsymbol{V}_n(D))}{\sum_{j=1}^m \sigma_j^2(\boldsymbol{V}_n(D))} + \kappa\, m, \quad (11)$$

with $\sigma_j(\boldsymbol{V}_n(D))$ the $j$th singular vector of $\boldsymbol{V}_n(D)$ and $\kappa$ a parameter. An alternative way of selecting the correct linear model (in feature space) for noisy data can be found in [11].

**Remark 3 (Suboptimality in the Stochastic Case).** Notice that, in the case of hyperplanes, the least-squares solution for $\boldsymbol{c}_n$ is obtained by minimizing $\|\boldsymbol{c}^T \boldsymbol{V}_n(D)\|^2$ subject to $\|\boldsymbol{c}_n\| = 1$. However, when $n > 1$ the so-found $\boldsymbol{c}_n$ does not minimize the sum of least-square errors $\sum_j \min_{i=1,\dots,n} (\boldsymbol{b}_i^T \boldsymbol{x}_j)^2$. Instead, it minimizes a "weighted version" of the least-square errors

$$\sum_j \alpha_j \min_{i=1,\dots,n} (\boldsymbol{b}_i^T \boldsymbol{x}_j)^2 \doteq \sum_j \prod_{i=1}^{n} (\boldsymbol{b}_i^T \boldsymbol{x}_j)^2 = \|\boldsymbol{c}^T \boldsymbol{V}_n(D)\|^2, \quad (12)$$

where the weight $\alpha_j$ is conveniently chosen so as to eliminate the minimization over $i = 1, \dots, n$. Such a "softening" of the objective function permits a global algebraic solution because the softened error does not depend on the membership of one point to one of the hyperplanes. This least-squares solution for $\boldsymbol{c}_n$ offers a suboptimal approximation for the original stochastic objective when the variance of the noise is small. This solution can be used to initialize other iterative optimization schemes (such as EM) to further minimize the original stochastic objective.

### 3.4 Obtaining a Basis and the Dimension of Each Subspace by Polynomial Differentiation

In this section, we show that one can obtain the bases $\{B_i\}_{i=1}^n$ for the complement of the $n$ subspaces and their dimensions $\{d_i\}_{i=1}^n$ by differentiating all the polynomials obtained from the left null space of the embedded data matrix $\boldsymbol{V}_n(D)$.

For the sake of simplicity, let us first consider the case of hyperplanes, i.e., subspaces of equal dimension $d_i = D - 1$, for $i = 1, \dots, n$. In this case, there is only one vector $\boldsymbol{b}_i \in \mathbb{R}^D$ normal to subspace $S_i$. Therefore, there is only one polynomial representing the $n$ hyperplanes, namely, $p_n(\boldsymbol{x}) = (\boldsymbol{b}_1^T \boldsymbol{x}) \cdots (\boldsymbol{b}_n^T \boldsymbol{x}) = \boldsymbol{c}_n^T \nu_n(\boldsymbol{x})$ and its coefficient vector $\boldsymbol{c}_n$ can be computed as the unique vector in the left null space of $\boldsymbol{V}_n(D)$. Consider now the derivative of $p_n(\boldsymbol{x})$

$$Dp_n(\boldsymbol{x}) = \frac{\partial p_n(\boldsymbol{x})}{\partial \boldsymbol{x}} = \frac{\partial}{\partial \boldsymbol{x}} \prod_{i=1}^{n} (\boldsymbol{b}_i^T \boldsymbol{x}) = \sum_{i=1}^{n} (\boldsymbol{b}_i) \prod_{\ell \neq i} (\boldsymbol{b}_\ell^T \boldsymbol{x}), \quad (13)$$

at a point $\boldsymbol{y}_i \in S_i$, i.e., $\boldsymbol{y}_i$ is such that $\boldsymbol{b}_i^T \boldsymbol{y}_i = 0$. Then, all terms in (13), except the $i$th, vanish, because $\prod_{\ell \neq i} (\boldsymbol{b}_\ell^T \boldsymbol{y}_j) = 0$ for $j \neq i$, so that we can immediately obtain the normal vectors as

$$\boldsymbol{b}_i = \frac{Dp_n(\boldsymbol{y}_i)}{\|Dp_n(\boldsymbol{y}_i)\|}, \quad i = 1, \dots, n. \quad (14)$$

Therefore, in a *semisupervised learning scenario* in which we are given only *one positive example per class*, the hyperplane segmentation problem can be solved *analytically* by evaluating the *derivatives* of $p_n(\boldsymbol{x})$ at the points with known labels.

As it turns out, the same principle applies to subspaces of arbitrary dimensions. This fact should come at no surprise. The zero set of each vanishing polynomial $p_{n\ell}$ is just a surface in $\mathbb{R}^D$; therefore, the derivative of $p_{n\ell}$ at a point $\boldsymbol{y}_i \in S_i$, $Dp_{n\ell}(\boldsymbol{y}_i)$, gives a vector normal to the surface. Since a union of subspaces is locally flat, i.e., in a neighborhood of $\boldsymbol{y}_i$ the surface is merely the subspace $S_i$, then the derivative at $\boldsymbol{y}_i$ lies in the orthogonal complement $S_i^\perp$ of $S_i$. By evaluating the derivatives of *all* the polynomials in $I_n$ at the same point $\boldsymbol{y}_i$, we obtain a set of normal vectors that span the orthogonal complement of $S_i$, as stated in Theorem 3. Fig. 2 illustrates the theorem for the case of a plane and a line described in Section 2.

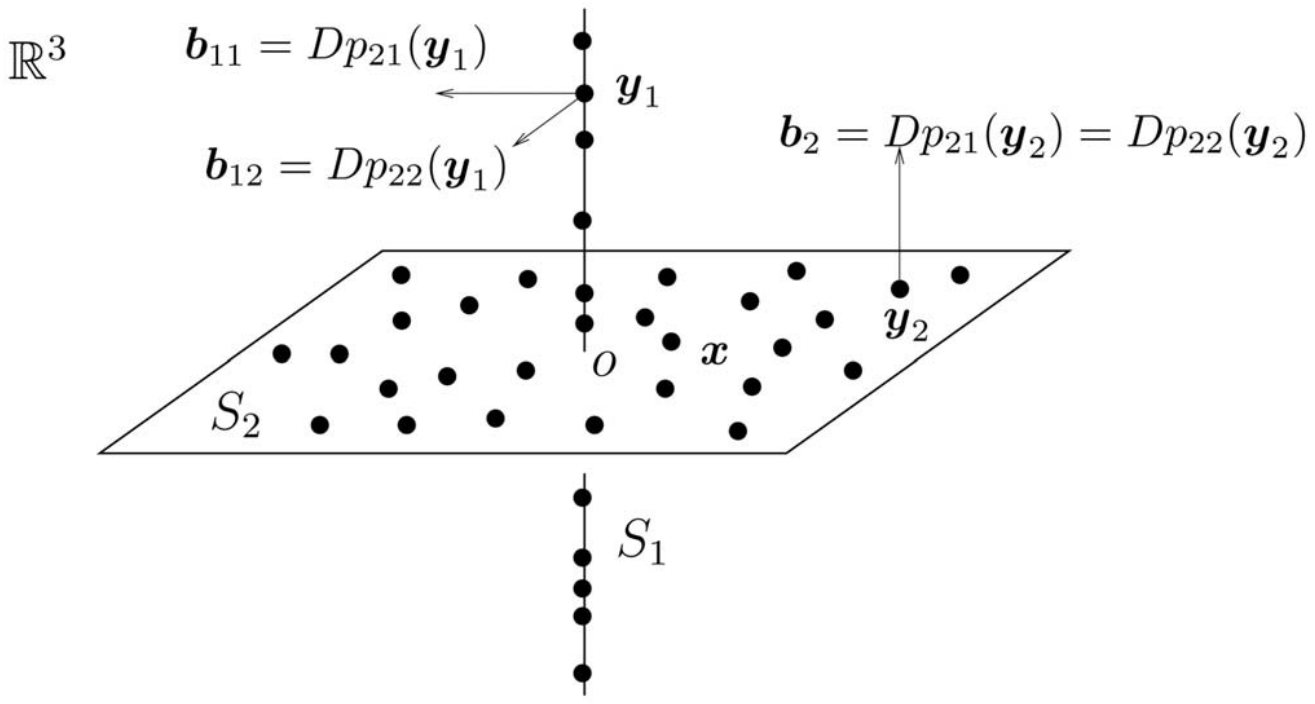

Fig. 2. The derivatives of the two polynomials $x_1 x_2$ and $x_1 x_3$ evaluated at a point $\boldsymbol{y}_1$ on the line $S_1$ give two normal vectors to the line. Similarly, the derivatives at a point $\boldsymbol{y}_2$ on the plane $S_2$ give the normal vector to the plane.

**Theorem 3 (Obtaining Subspace Bases and Dimensions by Polynomial Differentiation).** *Let $I_n$ be (the space of coefficient vectors of) the set of polynomials of degree $n$ that vanish on the $n$ subspaces. If the data set $\boldsymbol{X}$ is such that $\dim(\mathrm{null}(\boldsymbol{V}_n(D))) = \dim(I_n) = m_n$ and one point $\boldsymbol{y}_i \in S_i$ but $\boldsymbol{y}_i \notin S_j$ for $j \neq i$ is given for each subspace $S_i$, then we have*

$$S_i^\perp = \mathrm{span}\left\{ \frac{\partial}{\partial \boldsymbol{x}} \boldsymbol{c}_n^T \nu_n(\boldsymbol{x}) \Big|_{\boldsymbol{x}=\boldsymbol{y}_i}, \ \forall \boldsymbol{c}_n \in \mathrm{null}(\boldsymbol{V}_n(D)) \right\}. \quad (15)$$

*Therefore, the dimensions of the subspaces are given by*

$$d_i = D - \mathrm{rank}\big(DP_n(\boldsymbol{y}_i)\big) \quad \text{for} \quad i = 1, \dots, n, \quad (16)$$

*with $P_n(\boldsymbol{x}) = [p_{n1}(\boldsymbol{x}), \dots, p_{nm_n}(\boldsymbol{x})] \in \mathbb{R}^{1 \times m_n}$ and $DP_n(\boldsymbol{x}) = [Dp_{n1}(\boldsymbol{x}), \dots, Dp_{nm_n}(\boldsymbol{x})] \in \mathbb{R}^{D \times m_n}$.*

As a consequence of Theorem 3, we already have the sketch of an algorithm for segmenting subspaces of arbitrary dimensions in a semisupervised learning scenario in which we are given *one positive example per class* $\{\boldsymbol{y}_i \in S_i\}_{i=1}^n$:

1. Compute a basis for the left null space of $\boldsymbol{V}_n(D)$ using, for example, SVD.
2. Evaluate the derivatives of the polynomial $\boldsymbol{c}_n^T \nu_n(\boldsymbol{x})$ at $\boldsymbol{y}_i$ for each $\boldsymbol{c}_n$ in the basis of $\mathrm{null}(\boldsymbol{V}_n(D))$ to obtain a set of normal vectors in $S_i^\perp$.
3. Compute a basis $B_i$ for $S_i^\perp$ by applying PCA to the normal vectors obtained in Step 2. PCA automatically gives the dimension of each subspace $d_i = \dim(S_i)$.
4. Cluster the data by assigning point $\boldsymbol{x}_j$ to subspace $i$ if

$$i = \arg \min_{\ell=1,\dots,n} \|B_\ell^T \boldsymbol{x}_j\|. \quad (17)$$

**Remark 4 (Estimating the Bases from Noisy Data Points).** With a moderate level of noise in the data, we can still obtain a basis for each subspace and cluster the data as above. This is because we are applying PCA to the derivatives of the polynomials and both the coefficients of the polynomials and their derivatives depend continuously on the data.

Notice also that we can obtain the dimension of each subspace by looking at the singular values of the matrix of derivatives, similarly to (11).

**Remark 5 (Computing Derivatives of Homogeneous Polynomials).** Notice that given $\boldsymbol{c}_n$ the computation of the derivatives of $p_n(\boldsymbol{x}) = \boldsymbol{c}_n^T \nu_n(\boldsymbol{x})$ does *not* involve taking derivatives of the (possibly noisy) data. For instance, one may compute the derivatives as $\frac{\partial p_n(\boldsymbol{x})}{\partial x_k} = \boldsymbol{c}_n^T \frac{\nu_n(\boldsymbol{x})}{\partial x_k} = \boldsymbol{c}_n^T E_{nk} \nu_{n-1}(\boldsymbol{x})$, where $E_{nk} \in \mathbb{R}^{M_n(D) \times M_{n-1}(D)}$ is a constant matrix that depends on the exponents of the different monomials in the Veronese map $\nu_n(\boldsymbol{x})$.

### 3.5 Choosing One Point per Subspace by Polynomial Division

Theorem 3 demonstrates that one can obtain a basis for each $S_i^\perp$ directly from the derivatives of the polynomials representing the union of subspaces. However, in order to proceed we need to have one point per subspace, i.e., we need to know the vectors $\{\boldsymbol{y}_i\}_{i=1}^n$.

In this section, we show how to select these $n$ points in the *unsupervised learning scenario* in which we do not know the label for any of the data points. To this end, notice that we can always choose a point $\boldsymbol{y}_n$ lying on one of the subspaces, say $S_n$, by checking that $P_n(\boldsymbol{y}_n) = \boldsymbol{0}^T$. Since we are given a set of data points $\boldsymbol{X} = \{\boldsymbol{x}_j\}_{j=1}^n$ lying on the subspaces, in principle, we could choose $\boldsymbol{y}_n$ to be any of the data points. However, in the presence of noise and outliers, a random choice of $\boldsymbol{y}_n$ may be far from the true subspaces. In Section 2, we chose a point in the data set $\boldsymbol{X}$ that minimizes $\|P_n(\boldsymbol{x})\|$. However, such a choice has the following problems:

1. The value $\|P_n(\boldsymbol{x})\|$ is merely an *algebraic* error, i.e., it does not represent the *geometric* distance from $\boldsymbol{x}$ to its closest subspace. In principle, finding the geometric distance from $\boldsymbol{x}$ to its closest subspace is a difficult problem because we do not know the normal bases $\{B_i\}_{i=1}^n$.
2. Points $\boldsymbol{x}$ lying close to the intersection of two or more subspaces could be chosen. However, at a point $\boldsymbol{x}$ in the intersection of two or more subspaces, we often have $Dp_n(\boldsymbol{x}) = \boldsymbol{0}$. Thus, one should avoid choosing such points, as they give very noisy estimates of the normal vectors.

As it turns out, one can avoid both of these problems thanks to the following lemma:

**Lemma 1.** *Let $\tilde{\boldsymbol{x}}$ be the projection of $\boldsymbol{x} \in \mathbb{R}^D$ onto its closest subspace. The Euclidean distance from $\boldsymbol{x}$ to $\tilde{\boldsymbol{x}}$ is*

$$\|\boldsymbol{x} - \tilde{\boldsymbol{x}}\| = n\sqrt{P_n(\boldsymbol{x})\big(DP_n(\boldsymbol{x})^T DP_n(\boldsymbol{x})\big)^\dagger P_n(\boldsymbol{x})^T} + O\big(\|\boldsymbol{x} - \tilde{\boldsymbol{x}}\|^2\big),$$

*where $P_n(\boldsymbol{x}) = [p_{n1}(\boldsymbol{x}), \ldots, p_{nm_n}(\boldsymbol{x})] \in \mathbb{R}^{1\times m_n}$, $DP_n(\boldsymbol{x}) = [Dp_{n1}(\boldsymbol{x}), \ldots, Dp_{nm_n}(\boldsymbol{x})] \in \mathbb{R}^{D\times m_n}$, and $A^\dagger$ is the Moore-Penrose inverse of $A$.*

**Proof.** The projection $\tilde{\boldsymbol{x}}$ of a point $\boldsymbol{x}$ onto the zero set of the polynomials $\{p_{n\ell}\}_{\ell=1}^{m_n}$ can be obtained as the solution of the following constrained optimization problem

$$\begin{aligned} &\min && \|\tilde{\boldsymbol{x}} - \boldsymbol{x}\|^2 \\ &\text{subject to} && p_{n\ell}(\tilde{\boldsymbol{x}}) = 0 \quad \ell = 1, \ldots, m_n. \end{aligned} \tag{18}$$

By using Lagrange multipliers $\lambda \in \mathbb{R}^{m_n}$, we can convert this problem into the unconstrained optimization problem

$$\min_{\tilde{\boldsymbol{x}},\lambda} \|\tilde{\boldsymbol{x}} - \boldsymbol{x}\|^2 + P_n(\tilde{\boldsymbol{x}})\lambda. \tag{19}$$

From the first order conditions with respect to $\tilde{\boldsymbol{x}}$, we have $2(\tilde{\boldsymbol{x}} - \boldsymbol{x}) + DP_n(\tilde{\boldsymbol{x}})\lambda = \boldsymbol{0}$. After multiplying on the left by $(\tilde{\boldsymbol{x}} - \boldsymbol{x})^T$ and $(DP_n(\tilde{\boldsymbol{x}}))^T$, respectively, we obtain

$$\|\tilde{\boldsymbol{x}} - \boldsymbol{x}\|^2 = \frac{1}{2}\boldsymbol{x}^T DP_n(\tilde{\boldsymbol{x}})\lambda, \qquad \text{and} \tag{20}$$

$$\lambda = 2\big(DP_n(\tilde{\boldsymbol{x}})^T DP_n(\tilde{\boldsymbol{x}})\big)^\dagger DP_n(\tilde{\boldsymbol{x}})^T \boldsymbol{x}, \tag{21}$$

where we have used the fact that $(DP_n(\tilde{\boldsymbol{x}}))^T \tilde{\boldsymbol{x}} = nP_n(\tilde{\boldsymbol{x}}) = 0$ because $D\nu_n(\tilde{\boldsymbol{x}})^T \tilde{\boldsymbol{x}} = n\nu_n(\tilde{\boldsymbol{x}})$. After replacing (21) on (20), the squared distance from $\boldsymbol{x}$ to its closest subspace is given by

$$\|\tilde{\boldsymbol{x}} - \boldsymbol{x}\|^2 = \boldsymbol{x}^T DP_n(\tilde{\boldsymbol{x}})\big(DP_n(\tilde{\boldsymbol{x}})^T DP_n(\tilde{\boldsymbol{x}})\big)^\dagger DP_n(\tilde{\boldsymbol{x}})^T \boldsymbol{x}. \tag{22}$$

After expanding in Taylor series about $\tilde{\boldsymbol{x}} = \boldsymbol{x}$ and noticing that $DP_n(\boldsymbol{x})^T \boldsymbol{x} = nP_n(\boldsymbol{x})^T$, we obtain

$$\|\tilde{\boldsymbol{x}} - \boldsymbol{x}\|^2 \approx n^2 P_n(\boldsymbol{x})\big(DP_n(\boldsymbol{x})^T DP_n(\boldsymbol{x})\big)^\dagger P_n(\boldsymbol{x})^T, \tag{23}$$

which completes the proof. □

Thanks to Lemma 1, we can immediately choose a point $\boldsymbol{y}_n$ lying in (close to) one of the subspaces and not in (far from) the other subspaces as

$$\boldsymbol{y}_n = \arg\min_{\boldsymbol{x}\in\boldsymbol{X}: DP_n(\boldsymbol{x})\neq 0} P_n(\boldsymbol{x})\big(DP_n(\boldsymbol{x})^T DP_n(\boldsymbol{x})\big)^\dagger P_n(\boldsymbol{x})^T, \tag{24}$$

and then compute the basis $B_n \in \mathbb{R}^{D\times(D-d_n)}$ for $S_n^\perp$ by applying PCA to $DP_n(\boldsymbol{y}_n)$.

In order to find a point $\boldsymbol{y}_{n-1}$ lying in (close to) one of the remaining $(n-1)$ subspaces but not in (far from) $S_n$, we find a new set of polynomials $\{p_{(n-1)\ell}(\boldsymbol{x})\}$ defining the algebraic set $\cup_{i=1}^{n-1} S_i$. In the case of hyperplanes, there is only one such polynomial, namely,

$$p_{n-1}(\boldsymbol{x}) \doteq (\boldsymbol{b}_1 \boldsymbol{x}) \cdots (\boldsymbol{b}_{n-1}^T \boldsymbol{x}) = \frac{p_n(\boldsymbol{x})}{\boldsymbol{b}_n^T \boldsymbol{x}} = \boldsymbol{c}_{n-1}^T \nu_{n-1}(\boldsymbol{x}).$$

Therefore, we can obtain $p_{n-1}(\boldsymbol{x})$ by *polynomial division*. Notice that dividing $p_n(\boldsymbol{x})$ by $\boldsymbol{b}_n^T \boldsymbol{x}$ is a *linear problem* of the form $\boldsymbol{c}_{n-1}^T R_n(\boldsymbol{b}_n) = \boldsymbol{c}_n^T$, where $R_n(\boldsymbol{b}_n) \in \mathbb{R}^{M_{n-1}(D)\times M_n(D)}$. This is because solving for the coefficients of $p_{n-1}(\boldsymbol{x})$ is equivalent to solving the equations $(\boldsymbol{b}_n^T \boldsymbol{x})(\boldsymbol{c}_{n-1}^T \nu_n(\boldsymbol{x})) = \boldsymbol{c}_n^T \nu_n(\boldsymbol{x})$, where $\boldsymbol{b}_n$ and $\boldsymbol{c}_n$ are already known.

**Example 2.** If $n = 2$ and $\boldsymbol{b}_2 = [b_1, b_2, b_3]^T$, then the matrix $R_2(\boldsymbol{b}_2)$ is given by

$$R_2(\boldsymbol{b}_2) = \begin{bmatrix} b_1 & b_2 & b_3 & 0 & 0 & 0 \\ 0 & b_1 & 0 & b_2 & b_3 & 0 \\ 0 & 0 & b_1 & 0 & b_2 & b_3 \end{bmatrix} \in \mathbb{R}^{3\times 6}.$$

In the case of subspaces of varying dimensions, in principle, we cannot simply divide the entries of the polynomial vector $P_n(\boldsymbol{x})$ by $\boldsymbol{b}_n^T \boldsymbol{x}$ for any column $\boldsymbol{b}_n$ of $B_n$ because the polynomials $\{p_{n\ell}(\boldsymbol{x})\}$ may not be factorizable.[2] Furthermore, they do not necessarily have the common factor $\boldsymbol{b}_n^T\boldsymbol{x}$. The following theorem resolves this difficulty by showing how to compute the polynomials associated with the remaining subspaces $\cup_{i=1}^{n-1}S_i$:

**Theorem 4 (Obtaining Points by Polynomial Division).** *Let $I_n$ be (the space of coefficient vectors of) the set of polynomials vanishing on the $n$ subspaces. If the data set $\boldsymbol{X}$ is such that $\dim(\text{null}(\boldsymbol{V}_n(D))) = \dim(I_n)$, then the set of homogeneous polynomials of degree $(n-1)$ that vanish on the algebraic set $\cup_{i=1}^{n-1}S_i$ is spanned by $\{\boldsymbol{c}_{n-1}^T\nu_{n-1}(\boldsymbol{x})\}$, where the vectors of coefficients $\boldsymbol{c}_{n-1}\in\mathbb{R}^{M_{n-1}(D)}$ must satisfy*

$$\boldsymbol{c}_{n-1}^T R_n(\boldsymbol{b}_n)\boldsymbol{V}_n(D) = \boldsymbol{0}^T, \quad \text{for all} \quad \boldsymbol{b}_n \in S_n^\perp. \tag{25}$$

**Proof.** We first show the necessity. That is, any polynomial of degree $n-1$, $\boldsymbol{c}_{n-1}^T\nu_{n-1}(\boldsymbol{x})$, that vanishes on $\cup_{i=1}^{n-1}S_i$ satisfies the above equation. Since a point $\boldsymbol{x}$ in the original algebraic set $\cup_{i=1}^{n}S_i$ belongs to either $\cup_{i=1}^{n-1}S_i$ or $S_n$, we have $\boldsymbol{c}_{n-1}^T\nu_{n-1}(\boldsymbol{x}) = 0$ or $\boldsymbol{b}_n^T\boldsymbol{x} = 0$ for all $\boldsymbol{b}_n \in S_n^\perp$. Hence, $p_n(\boldsymbol{x}) \doteq (\boldsymbol{c}_{n-1}^T\nu_{n-1}(\boldsymbol{x}))(\boldsymbol{b}_n^T\boldsymbol{x}) = 0$. If we denote $p_n(\boldsymbol{x})$ as $\boldsymbol{c}_n^T\nu_n(\boldsymbol{x})$, then the coefficient vector $\boldsymbol{c}_n$ must be in $\text{null}(\boldsymbol{V}_n(D))$. From $\boldsymbol{c}_n^T\nu_n(\boldsymbol{x}) = (\boldsymbol{c}_{n-1}^T\nu_{n-1}(\boldsymbol{x}))(\boldsymbol{b}_n^T\boldsymbol{x})$, the relationship between $\boldsymbol{c}_n$ and $\boldsymbol{c}_{n-1}$ can be written as $\boldsymbol{c}_{n-1}^T R_n(\boldsymbol{b}_n) = \boldsymbol{c}_n^T$. Since $\boldsymbol{c}_n^T\boldsymbol{V}_n(D) = \boldsymbol{0}^T$, $\boldsymbol{c}_{n-1}$ needs to satisfy the following linear system of equations $\boldsymbol{c}_{n-1}^T R_n(\boldsymbol{b}_n)\boldsymbol{V}_n(D) = \boldsymbol{0}^T$.

We now show the sufficiency. That is, if $\boldsymbol{c}_{n-1}$ is a solution to (25), then for all $\boldsymbol{b}_n \in S_n^\perp$, $\boldsymbol{c}_n^T = \boldsymbol{c}_{n-1}^T R_n(\boldsymbol{b}_n)$ is in $\text{null}(\boldsymbol{V}_n(D))$. From the construction of $R_n(\boldsymbol{b}_n)$, we have $\boldsymbol{c}_n^T\nu_n(\boldsymbol{x}) = (\boldsymbol{c}_{n-1}^T\nu_{n-1}(\boldsymbol{x}))(\boldsymbol{b}_n^T\boldsymbol{x})$. Then, for every $\boldsymbol{x} \in \cup_{i=1}^{n-1}S_i$ but not in $S_n$, we have $\boldsymbol{c}_{n-1}^T\nu_{n-1}(\boldsymbol{x}) = 0$ because there is a $\boldsymbol{b}_n$ such that $\boldsymbol{b}_n^T\boldsymbol{x} \neq 0$. Therefore, $\boldsymbol{c}_{n-1}^T\nu_{n-1}(\boldsymbol{x})$ is a homogeneous polynomial of degree $(n-1)$ that vanishes on $\cup_{i=1}^{n-1}S_i$. □

Thanks to Theorem 4, we can obtain a collection of polynomials $\{p_{(n-1)\ell}(\boldsymbol{x})\}_{\ell=1}^{m_{n-1}}$ representing $\cup_{i=1}^{n-1}S_i$ from the intersection of the left null spaces of $R_n(\boldsymbol{b}_n)\boldsymbol{V}_n(D) \in \mathbb{R}^{M_{n-1}(D)\times N}$ for all $\boldsymbol{b}_n \in S_n^\perp$. We can then repeat the same procedure to find a basis for the remaining subspaces. We thus obtain the following *Generalized Principal Component Analysis* (GPCA) algorithm (Algorithm 1) for segmenting $n$ subspaces of unknown and possibly different dimensions.

**Algorithm 1**
**(GPCA: Generalized Principal Component Analysis)**
**set** $\boldsymbol{V}_n = [\nu_n(\boldsymbol{x}_1),\ldots,\nu_n(\boldsymbol{x}_N)] \in \mathbb{R}^{M_n(D)\times N}$;
**for** $i = n:1$ **do**
  **solve** $\boldsymbol{c}^T\boldsymbol{V}_i = 0$ to obtain a basis $\{\boldsymbol{c}_{i\ell}\}_{\ell=1}^{m_i}$ of $\text{null}(\boldsymbol{V}_i)$, where the number of polynomials $m_i$ is obtained as in (11);
  **set** $P_i(\boldsymbol{x}) = [p_{i1}(\boldsymbol{x}),\ldots,p_{im_i}(\boldsymbol{x})] \in \mathbb{R}^{1\times m_i}$, where $p_{i\ell}(\boldsymbol{x}) = \boldsymbol{c}_{i\ell}^T\nu_i(\boldsymbol{x})$ for $\ell = 1,\ldots,m_i$;
  **do**

$$\boldsymbol{y}_i = \arg\min_{\boldsymbol{x}\in\boldsymbol{X}:DP_i(\boldsymbol{x})\neq 0} P_i(\boldsymbol{x})\big(DP_i(\boldsymbol{x})^T DP_i(\boldsymbol{x})\big)^\dagger P_i(\boldsymbol{x})^T,$$

$$B_i = PCA\big(DP_i(\boldsymbol{y}_i)\big),$$

$$\boldsymbol{V}_{i-1} = \big[R_i(\boldsymbol{b}_{i1})\boldsymbol{V}_i,\ldots,R_i(\boldsymbol{b}_{i,D-d_i})\boldsymbol{V}_i\big], \text{with}$$
  $\boldsymbol{b}_{ij}$ columns of $B_i$;
  **end do**
**end for**
**for** $j = 1:N$ **do**
  assign point $\boldsymbol{x}_j$ to subspace $S_i$ if $i = \arg\min_\ell \|B_\ell^T\boldsymbol{x}_j\|$;
**end for**

**Remark 6 (Avoiding Polynomial Division).** Notice that one may avoid computing $P_i$ for $i < n$ by using a heuristic distance function to choose the points $\{\boldsymbol{y}_i\}_{i=1}^n$. Since a point in $\cup_{\ell=i}^n S_\ell$ must satisfy $\|B_i^T\boldsymbol{x}\|\cdots\|B_n^T\boldsymbol{x}\| = 0$, we can choose a point $\boldsymbol{y}_{i-1}$ on $\cup_{\ell=1}^{i-1}S_\ell$ as:

$$\boldsymbol{y}_{i-1} = \arg\min_{\boldsymbol{x}\in\boldsymbol{X}:DP_n(\boldsymbol{x})\neq 0} \frac{\sqrt{P_n(\boldsymbol{x})(DP_n(\boldsymbol{x})^T DP_n(\boldsymbol{x}))^\dagger P_n(\boldsymbol{x})^T} + \delta}{\|B_i^T\boldsymbol{x}\|\cdots\|B_n^T\boldsymbol{x}\| + \delta},$$

where a small number $\delta > 0$ is chosen to avoid cases in which both the numerator and the denominator are zero (e.g., with perfect data).

**Remark 7 (Robustness and Outlier Rejection).** In practice, there could be points in $\boldsymbol{X}$ that are far away from any of the subspaces, i.e., outliers. By detecting and rejecting outliers, we can typically ensure a much better estimate of the subspaces. Many methods from robust statistics can be deployed to detect and reject the outliers [5], [11]. For instance, the function

$$d^2(\boldsymbol{x}) = P_n(\boldsymbol{x})\big(DP_n(\boldsymbol{x})^T DP_n(\boldsymbol{x})\big)^\dagger P_n(\boldsymbol{x})^T$$

approximates the squared distance of a point $\boldsymbol{x}$ to the subspaces. From the $d^2$-histogram of the sample set $\boldsymbol{X}$, we may exclude from $\boldsymbol{X}$ all points that have unusually large $d^2$ values and use only the remaining sample points to re-estimate the polynomials before computing the normals. For instance, if we assume that the sample points are drawn around each subspace from independent Gaussian distributions with a small variance $\sigma^2$, then $\frac{d^2}{\sigma^2}$ is approximately a $\chi^2$-distribution with $\sum_i(D-d_i)$ degrees of freedom. We can apply standard $\chi^2$-test to reject samples which deviate significantly from this distribution. Alternatively, one can detect and reject outliers using Random Sample Consensus (RANSAC) [5]. One can choose $M_n(D)$ data points at random, estimate a collection of polynomials passing through those points, determine their degree of support among the other points, and then choose the set of polynomials giving a large degree of support. This method is expected to be effective when $M_n(D)$ is small. An open problem is how to combine GPCA with methods from robust statistics in order to improve the robustness of GPCA to outliers.

## 4 Extensions to the Basic GPCA Algorithm

In this section, we discuss some extensions of GPCA that deal with practical situations such as low-dimensional subspaces of a high-dimensional space and unknown number of subspaces.

### 4.1 Projection and Minimum Representation

When the dimension of the ambient space $D$ is large, the complexity of GPCA becomes prohibitive because $M_n(D)$ is of the order $n^D$. However, in most practical situations, we are interested in modeling the data as a union of subspaces

2. Recall that we can only compute a basis for the null space of $\boldsymbol{V}_n(D)$, and that linear combinations of factorizable polynomials are not necessarily factorizable. For example, $x_1^2 + x_1x_2$ and $x_2^2 - x_1x_2$ are both factorizable, but their sum $x_1^2 + x_2^2$ is not.

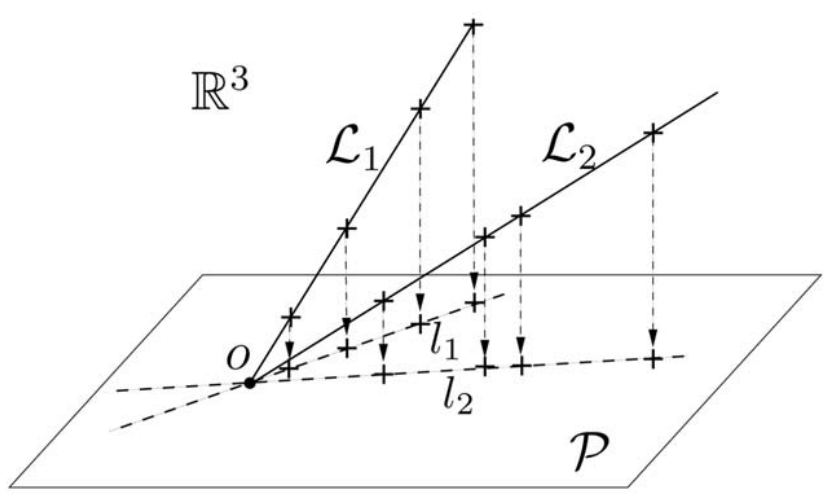

Fig. 3. A linear projection of two one-dimensional subspaces $\mathcal{L}_1, \mathcal{L}_2$ in $\mathbb{R}^3$ onto a two-dimensional plane $\mathcal{P}$ preserves the membership of each sample and the dimension of the lines.

of relatively small dimensions $\{d_i \ll D\}$. In such cases, it seems rather redundant to use $\mathbb{R}^D$ to represent such a low-dimensional linear structure. One way of reducing the dimensionality is to linearly project the data onto a lower-dimensional (sub)space. An example is shown in Fig. 3, where two lines $\mathcal{L}_1$ and $\mathcal{L}_2$ in $\mathbb{R}^3$ are projected onto a plane $\mathcal{P}$. In this case, segmenting the two lines in the three-dimensional space $\mathbb{R}^3$ is equivalent to segmenting the two projected lines $l_1$ and $l_2$ in the plane $\mathcal{P}$.

In general, we will distinguish between two different kinds of linear "projections." The first kind corresponds to the case in which the span of all the subspaces is a proper subspace of the ambient space, i.e., $\mathrm{span}(\cup_{i=1}^n S_i) \subset \mathbb{R}^D$. In this case, one may simply apply the classic PCA algorithm to the original data to eliminate the redundant dimensions. The second kind corresponds to the case in which the largest dimension of the subspaces, denoted by $d_{\max}$, is strictly less than $D-1$. When $d_{\max}$ is known, one may choose a $(d_{\max}+1)$-dimensional subspace $\mathcal{P}$ such that, by projecting onto this subspace:

$$\pi_{\mathcal{P}}: \ \boldsymbol{x} \in \mathbb{R}^D \quad \mapsto \quad \boldsymbol{x}' = \pi_{\mathcal{P}}(\boldsymbol{x}) \in \mathcal{P},$$

the dimension of each original subspace $S_i$ is preserved,[3] and the number of subspaces is preserved,[4] as stated in the following theorem:

**Theorem 5 (Segmentation-Preserving Projections).** *If a set of vectors $\{\boldsymbol{x}_j\}$ lie in $n$ subspaces of dimensions $\{d_i\}_{i=1}^n$ in $\mathbb{R}^D$ and if $\pi_{\mathcal{P}}$ is a linear projection into a subspace $\mathcal{P}$ of dimension $D'$, then the points $\{\pi_{\mathcal{P}}(\boldsymbol{x}_j)\}$ lie in $n' \le n$ linear subspaces of $\mathcal{P}$ of dimensions $\{d_i' \le d_i\}_{i=1}^n$. Furthermore, if $D > D' > d_{\max}$, then there is an open and dense set of projections that preserve the number and dimensions of the subspaces, i.e., $n' = n$ and $d_i' = d_i$ for $i = 1, \ldots, n$.*

Thanks to Theorem 5, if we are given a data set $\boldsymbol{X}$ drawn from a union of low-dimensional subspaces of a high-dimensional space, we can cluster the data set by first projecting $\boldsymbol{X}$ onto a generic subspace of dimension $D' = d_{\max}+1$ and then applying GPCA to the projected subspaces, as illustrated with the following sequence of steps:

$$\boldsymbol{X} \xrightarrow{\pi_{\mathcal{P}}} \boldsymbol{X}' \xrightarrow{\text{GPCA}} \cup_{i=1}^n \pi_{\mathcal{P}}(S_i) \xrightarrow{\pi_{\mathcal{P}}^{-1}} \cup_{i=1}^n S_i.$$

However, even though we have shown that the set of $(d_{\max}+1)$-dimensional subspaces $\mathcal{P} \subset \mathbb{R}^D$ that preserve the number and dimensions of the subspaces is an open and dense set, it remains unclear what a "good" choice for $\mathcal{P}$ is, especially when there is noise in the data. In practice, one may simply select a few random projections and choose the one that results in the smallest fitting error. Another alternative is to apply classic PCA to project onto a $(d_{\max}+1)$-dimensional affine subspace. The reader may refer to [1] for alternative ways of choosing a projection.

### 4.2 Identifying an Unknown Number of Subspaces of Unknown Dimensions

The solution to the subspace segmentation problem proposed in Section 3 assumes prior knowledge of the number of subspaces $n$. In practice, however, the number of subspaces $n$ may not be known beforehand, hence, we cannot estimate the polynomials representing the subspaces directly.

For the sake of simplicity, let us first consider the problem of determining the number of subspaces from a generic data set lying in a union of $n$ different hyperplanes $S_i = \{\boldsymbol{x} : \boldsymbol{b}_i^T \boldsymbol{x} = 0\}$. From Section 3, we know that in this case there is a *unique* polynomial of degree $n$ that vanishes in $Z = \cup_{i=1}^n S_i$, namely, $p_n(\boldsymbol{x}) = (\boldsymbol{b}_1^T \boldsymbol{x}) \cdots (\boldsymbol{b}_n^T \boldsymbol{x}) = \boldsymbol{c}_n^T \nu_n(\boldsymbol{x})$ and that its coefficient vector $\boldsymbol{c}_n$ lives in the left null space of the embedded data matrix $\boldsymbol{V}_n(D)$ defined in (9), hence, $\mathrm{rank}(\boldsymbol{V}_n) = M_n(D) - 1$. Clearly, there cannot be a polynomial of degree $i < n$ that vanishes in $Z$; otherwise, the data would lie in a union of $i < n$ hyperplanes. This implies that $\boldsymbol{V}_i(D)$ must be full rank for all $i < n$. In addition, notice that there is more than one polynomial of degree $i > n$ that vanishes on $Z$, namely, any multiple of $p_n$, hence, $\mathrm{rank}(\boldsymbol{V}_i(D)) < M_i(D) - 1$ if $i > n$. Therefore, the number of hyperplanes can be determined as the minimum degree such that the embedded data matrix drops rank, i.e.,

$$n = \min\{i : \mathrm{rank}(\boldsymbol{V}_i(D)) < M_i(D)\}. \qquad (26)$$

Consider now the case of data lying in subspaces of equal dimension $d_1 = d_2 = \cdots d_n = d < D-1$. For example, consider a set of points $\boldsymbol{X} = \{\boldsymbol{x}_i\}$ lying in two lines in $\mathbb{R}^3$, say,

$$S_1 = \{\boldsymbol{x} : x_2 = x_3 = 0\} \quad \text{and} \quad S_2 = \{\boldsymbol{x} : x_1 = x_3 = 0\}. \qquad (27)$$

If we construct the matrix of embedded data points $\boldsymbol{V}_n(D)$ for $n = 1$, we obtain $\mathrm{rank}(\boldsymbol{V}_1(3)) = 2 < 3$ because all the points lie also in the plane $x_3 = 0$. Therefore, we cannot determine the number of subspaces as in (26) because we would obtain $n = 1$, which is not correct. In order to determine the correct number of subspaces, recall from Section 4.1 that a linear projection onto a generic $(d+1)$-dimensional subspace $\mathcal{P}$ preserves the number and dimensions of the subspaces. Therefore, if we project the data onto $\mathcal{P}$, then the projected data lies in a union of $n$ hyperplanes of $\mathbb{R}^{d+1}$. By applying (26) to the projected data, we can obtain the number of subspaces from the embedded (projected) data matrix $\boldsymbol{V}_i(d+1)$ as

$$n = \min\{i : \mathrm{rank}(\boldsymbol{V}_i(d+1)) < M_i(d+1)\}. \qquad (28)$$

Of course, in order to apply this projection, we need to know the common dimension $d$ of all the subspaces. Clearly, if we project onto a subspace of dimension $\ell + 1 < d + 1$, then the number and dimension of the subspaces are no longer preserved. In fact, the projected data points lie in one subspace of dimension $\ell + 1$, and $\boldsymbol{V}_i(\ell+1)$ is of full rank for all $i$ (as long as $M_i(D) < N$). Therefore, we can determine the dimension of the subspaces as the minimum integer $\ell$ such that there is a degree $i$ for which $\boldsymbol{V}_i(\ell+1)$ drops rank, that is,

3. This requires that $\mathcal{P}$ be transversal to each $S_i^\perp$, i.e., $\mathrm{span}\{\mathcal{P}, S_i^\perp\} = \mathbb{R}^D$ for every $i = 1, \ldots, n$. Since $n$ is finite, this transversality condition can be easily satisfied. Furthermore, the set of positions for $\mathcal{P}$ which violate the transversality condition is only a zero-measure closed set [9].

4. This requires that all $\pi_{\mathcal{P}}(S_i)$ be transversal to each other in $\mathcal{P}$, which is guaranteed if we require $\mathcal{P}$ to be transversal to $S_i^\perp \cap S_j^\perp$ for $i, j = 1, .., n$. All $\mathcal{P}$s which violate this condition form again only a zero-measure set.

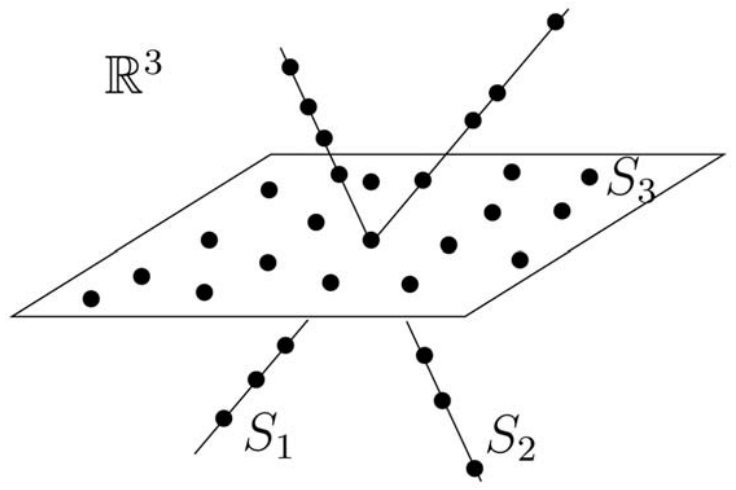

Fig. 4. A set of samples that can be interpreted as coming either from two lines and one plane or from two planes.

$$d = \min\{\ell : \exists\, i \geq 1 \text{ such } \operatorname{rank}(\boldsymbol{V}_i(\ell+1)) < M_i(\ell+1)\}. \quad (29)$$

In summary, when the subspaces are of equal dimension $d$, both the number of subspaces $n$ and their common dimension $d$ can be retrieved from (28) and (29) and the subspace segmentation problem can be subsequently solved by first projecting the data onto a $(d+1)$-dimensional subspace and then applying GPCA (Algorithm 1) to the projected data points.

**Remark 8.** In the presence of noise, one may not be able to estimate $d$ and $n$ from (29) and (28), respectively, because the matrix $\boldsymbol{V}_i(\ell+1)$ may be of full rank for all $i$ and $\ell$. Similarly to Remark 2, one can use model selection techniques to determine the rank of $\boldsymbol{V}_i(\ell)$. However, in practice this requires searching for up to possibly $(D-1)$ values for $d$ and $\lceil N/(D-1)\rceil$ values for $n$. One may refer to [11] for a more detailed discussion on selecting the best multiple-subspace model from noisy data, using model-selection criteria such as MML, MDL, AIC, and BIC.

Unfortunately, the situation is not so simple for subspaces of different dimensions. For instance, imagine that in addition to the two lines $S_1$ and $S_2$ we are also given data points on a plane $S_3 = \{\boldsymbol{x} : x_1 + x_2 = 0\}$, so that the overall configuration is similar to that shown in Fig. 4. In this case, we have $\operatorname{rank}(\boldsymbol{V}_1(3)) = 3 \not< 3$, $\operatorname{rank}(\boldsymbol{V}_2(3)) = 5 < 6$, and $\operatorname{rank}(\boldsymbol{V}_3(3)) = 6 < 10$. Therefore, if we try to determine the number of subspaces as the degree of the embedding for which the embedded data matrix drops rank we would obtain $n = 2$, which is incorrect again. The reason for this is clear: We can fit the data either with one polynomial of degree $n = 2$, which corresponds to the plane $S_3$ and the plane $\mathcal{P}$ spanned by the two lines, or with four polynomials of degree $n = 3$, which vanish precisely on the two lines $S_1$, $S_2$, and the plane $S_3$.

To resolve the difficulty in simultaneously determining the number and dimension of the subspaces, notice that the *algebraic set* $Z = \cup_{j=1}^n S_j$ can be decomposed into irreducible subsets $S_j$s—an irreducible algebraic set is also called a *variety*—and that the decomposition of $Z$ into $\{S_j\}_{j=1}^n$ is always unique [8]. Therefore, as long as we are able to correctly determine from the given sample points the underlying algebraic set $Z$ or the associated radical ideal $\mathcal{I}(Z)$,[5] in principle, the number of subspaces $n$ and their dimensions $\{d_j\}_{j=1}^n$ can always be uniquely determined in a purely algebraic fashion. In Fig. 4, for instance, the first interpretation (2 lines and 1 plane) would be the right one and the second one (two planes) would be incorrect because the two lines, which span one of the planes, are not an irreducible algebraic set.

Having established that the problem of subspace segmentation is equivalent to decomposing the algebraic ideal associated with the subspaces, we are left with deriving a computable scheme to achieve the goal of decomposing algebraic sets into varieties. To this end, notice that the set of all homogeneous polynomials that vanish in $Z$ can be graded by degree as

$$\mathcal{I}(Z) = \mathcal{I}_m \oplus \mathcal{I}_{m+1} \oplus \cdots \oplus \mathcal{I}_n \oplus \cdots, \quad (30)$$

where $m \leq n$ is the degree of the polynomial of minimum degree that fits all the data points. For each degree $i \geq m$, we can evaluate the derivatives of the polynomials in $\mathcal{I}_i$ at points in subspace $S_j$ and denote the collection of derivatives as

$$D_{i,j} \doteq \operatorname{span}\{\cup_{\boldsymbol{x}\in S_j}\{\nabla f\,|_{\boldsymbol{x}},\ \forall f \in \mathcal{I}_i\}\}, \quad j = 1, 2, \ldots, n. \quad (31)$$

Obviously, we have the following relationship:

$$D_{i,j} \subseteq D_{i+1,j} \subseteq S_j^{\perp}, \quad \forall i \geq m. \quad (32)$$

Therefore, for each degree $i \geq m$, we may compute a union of up to $n$ subspaces,

$$Z_i \doteq D_{i,1}^{\perp} \cup D_{i,2}^{\perp} \cup \cdots \cup D_{i,n}^{\perp} \supseteq Z, \quad (33)$$

which contains the original $n$ subspaces. Therefore, we can further partition $Z_i$ to obtain the original subspaces. More specifically, in order to segment an unknown number of subspaces of unknown and possibly different dimensions, we can first search for the minimum degree $i$ and dimension $\ell$ such that $\boldsymbol{V}_i(\ell+1)$ drops rank. In our example in Fig. 4, we obtain $i = 2$ and $\ell = 2$. By applying GPCA to the data set projected onto an $(\ell+1)$-dimensional space, we partition the data into up to $n$ subspaces $Z_i$ which contain the original $n$ subspaces. In our example, we partition the data into two planes $\mathcal{P}$ and $S_3$. Once these subspaces have been estimated, we can reapply the same process to each reducible subspace. In our example, the plane $\mathcal{P}$ will be separated into two lines $S_1$ and $S_2$, while the plane $S_3$ will remain unchanged. This recursive process stops when every subspace obtained can no longer be separated into lower-dimensional subspaces, or when a prespecified maximum number of subspaces $n_{\max}$ has been reached.

We summarize the above derivation with the recursive GPCA algorithm (Algorithm 2).

**Algorithm 2 Recursive GPCA Algorithm**

$n = 1$;
**repeat**
  build a data matrix $\boldsymbol{V}_n(D) \doteq [\nu_n(\boldsymbol{x}_1), \ldots, \nu_n(\boldsymbol{x}_N)] \in \mathbb{R}^{M_n(D)\times N}$ via the Veronese map $\nu_n$ of degree $n$;
  **if** $\operatorname{rank}(\boldsymbol{V}_n(D)) < M_n(D)$ **then**
    compute the basis $\{\boldsymbol{c}_{n\ell}\}$ of the left null space of $\boldsymbol{V}_n(D)$;
    obtain polynomials $\{p_{n\ell}(\boldsymbol{x}) \doteq \boldsymbol{c}_{n\ell}^T \nu_n(\boldsymbol{x})\}$;
    $\mathbf{Y} = \emptyset$;
    **for** $j = 1 : n$ **do**
      select a point $\boldsymbol{x}_j$ from $\boldsymbol{X} \setminus \mathbf{Y}$ (similar to Algorithm 1);
      obtain the subspace $S_j^{\perp}$ spanned by the derivatives $\operatorname{span}\{Dp_{n\ell}(\boldsymbol{x}_j)\}$; find the subset of points $\boldsymbol{X}_j \subset \boldsymbol{X}$ that belong to the subspace $S_j$; $\mathbf{Y} \leftarrow \mathbf{Y} \cup \boldsymbol{X}_j$;
      **Recursive-GPCA**($\boldsymbol{X}_j$); (with $S_j$ now as the ambient space)
    **end for**
    $n \leftarrow n_{max}$;
  **else**
    $n \leftarrow n + 1$;
  **end if**
**until** $n \geq n_{max}$.

5. The ideal of an algebraic set $Z$ is the set of all polynomials that vanish in $Z$. An ideal $\mathcal{I}$ is called radical if $f \in \mathcal{I}$ whenever $f^s \in \mathcal{I}$ for some integer $s$.

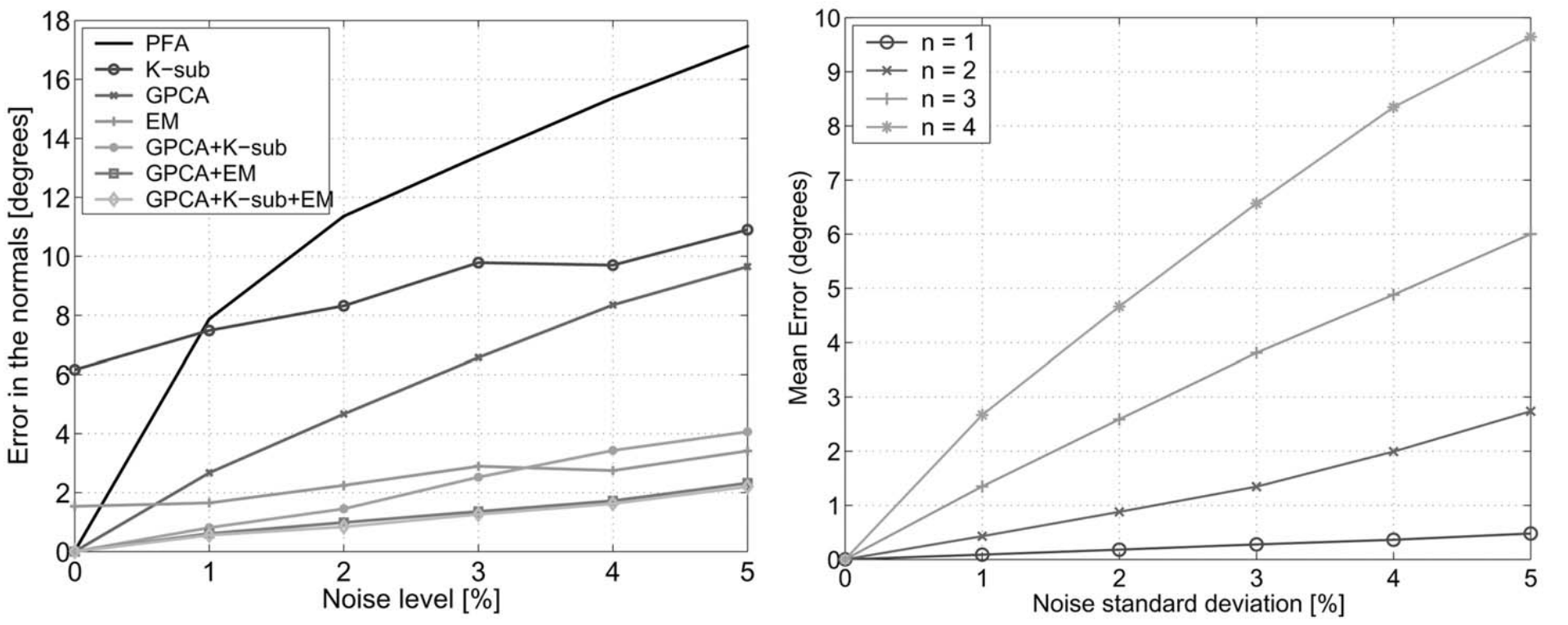

(a) (b)

Fig. 5. Error versus noise for data lying on two-dimensional subspaces of $\mathbb{R}^3$. (a) Error versus noise for $n = 4$. A comparison of PFA, GPCA ($\delta = 0.02$), K-subspaces and EM randomly initialized, K-subspaces and EM initialized with GPCA, and EM initialized with K-subspaces initialized with GPCA for $n = 4$ subspaces. (b) Error versus noise for $n = 1, \ldots, 4$. GPCA for $n = 1, \ldots, 4$ subspaces.

## 5 Experimental Results and Applications in Computer Vision

In this section, we first evaluate the performance of GPCA on synthetically generated data by comparing and combining it with the following approaches:

1. *Polynomial Factorization Algorithm (PFA)*. This algorithm is only applicable to the case of hyperplanes. It computes the normal vectors $\{\boldsymbol{b}_i\}_{i=1}^n$ to the $n$ hyperplanes by factorizing the homogeneous polynomial $p_n(\boldsymbol{x}) = (\boldsymbol{b}_1^T\boldsymbol{x})(\boldsymbol{b}_2^T\boldsymbol{x})\cdots(\boldsymbol{b}_n^T\boldsymbol{x})$ into a product of linear factors. See [24] for further details.
2. *K-subspaces*. Given an initial estimate for the subspace bases, this algorithm alternates between clustering the data points using the distance residual to the different subspaces and computing a basis for each subspace using standard PCA. See [10] for further details.
3. *Expectation Maximization (EM)*. This algorithm assumes that the data is corrupted with zero-mean Gaussian noise in the directions orthogonal to the subspace. Given an initial estimate for the subspace bases, EM alternates between clustering the data points (E-step) and computing a basis for each subspace (M-step) by maximizing the log-likelihood of the corresponding probabilistic model. See [19] for further details.

We then apply GPCA to various problems in computer vision such as face clustering under varying illumination, temporal video segmentation, two-view segmentation of linear motions, and multiview segmentation of rigid-body motions. However, it is *not* our intention to convince the reader that the proposed GPCA algorithm offers an optimal solution to each of these problems. In fact, one can easily obtain better segmentation results by using algorithms/systems specially designed for each of these tasks. We merely wish to point out that GPCA provides an effective tool to automatically detect the multiple-subspace structure present in these data sets in a noniterative fashion and that it provides a good initial estimate for any iterative algorithm.

### 5.1 Experiments on Synthetic Data

The experimental setup consists of choosing $n = 2, 3, 4$ collections of $N = 200n$ points in randomly chosen planes in $\mathbb{R}^3$. Zero-mean Gaussian noise with s.t.d. $\sigma$ from 0 percent to 5 percent along the subspace normals is added to the sample points. We run 1,000 trials for each noise level. For each trial, the error between the true (unit) normal vectors $\{\boldsymbol{b}_i\}_{i=1}^n$ and their estimates $\{\hat{\boldsymbol{b}}_i\}_{i=1}^n$ is computed as the mean angle between the normal vectors:

$$\text{error} \doteq \frac{1}{n}\sum_{i=1}^{n} \text{acos}\Big(\boldsymbol{b}_i^T\hat{\boldsymbol{b}}_i\Big)\,(\text{degrees}). \qquad (34)$$

Fig. 5a plots the mean error as a function of noise for $n = 4$. Similar results were obtained for $n = 2, 3$, though with smaller errors. Notice that the estimates of GPCA with the choice of $\delta = 0.02$ (see Remark 6) have an error that is only about 50 percent the error of the PFA. This is because GPCA deals automatically with noisy data by choosing the points $\{\boldsymbol{y}_i\}_{i=1}^n$ in an optimal fashion. The choice of $\delta$ was not important (results were similar for $\delta \in [0.001, 0.1]$). Notice also that both the K-subspaces and EM algorithms have a nonzero error in the noiseless case, showing that they frequently converge to a local minimum when a single randomly chosen initialization is used. When initialized with GPCA, both the K-subspaces and EM algorithms reduce the error to approximately 35-50 percent with respect to random initialization. The best performance is achieved by using GPCA to initialize the K-subspaces and EM algorithms.

Fig. 5b plots the estimation error of GPCA as a function of the number of subspaces $n$, for different levels of noise. As expected, the error increases rapidly as a function of $n$ because GPCA needs a minimum of $O(n^2)$ data points to linearly estimate the polynomials (see Section 4.1).

TABLE 1
Mean Computing Time and Mean Number of Iterations for Various Subspace Segmentation Algorithms

| Algorithms | $\boldsymbol{c}^T\boldsymbol{V}_n = \boldsymbol{0}^T$ | PFA | GPCA | K-sub |
|---|---|---|---|---|
| Time (sec.) | 0.0854 | 0.1025 | 0.1818 | 0.4637 |
| # Iterations | none | none | none | 19.7 |

| Algorithms | GPCA +K-sub | EM | GPCA+EM | GPCA+K-sub+EM |
|---|---|---|---|---|
| Time (sec.) | 0.2525 | 1.0408 | 0.6636 | 0.7528 |
| # Iterations | 7.1 | 30.8 | 17.1 | 15.0 |

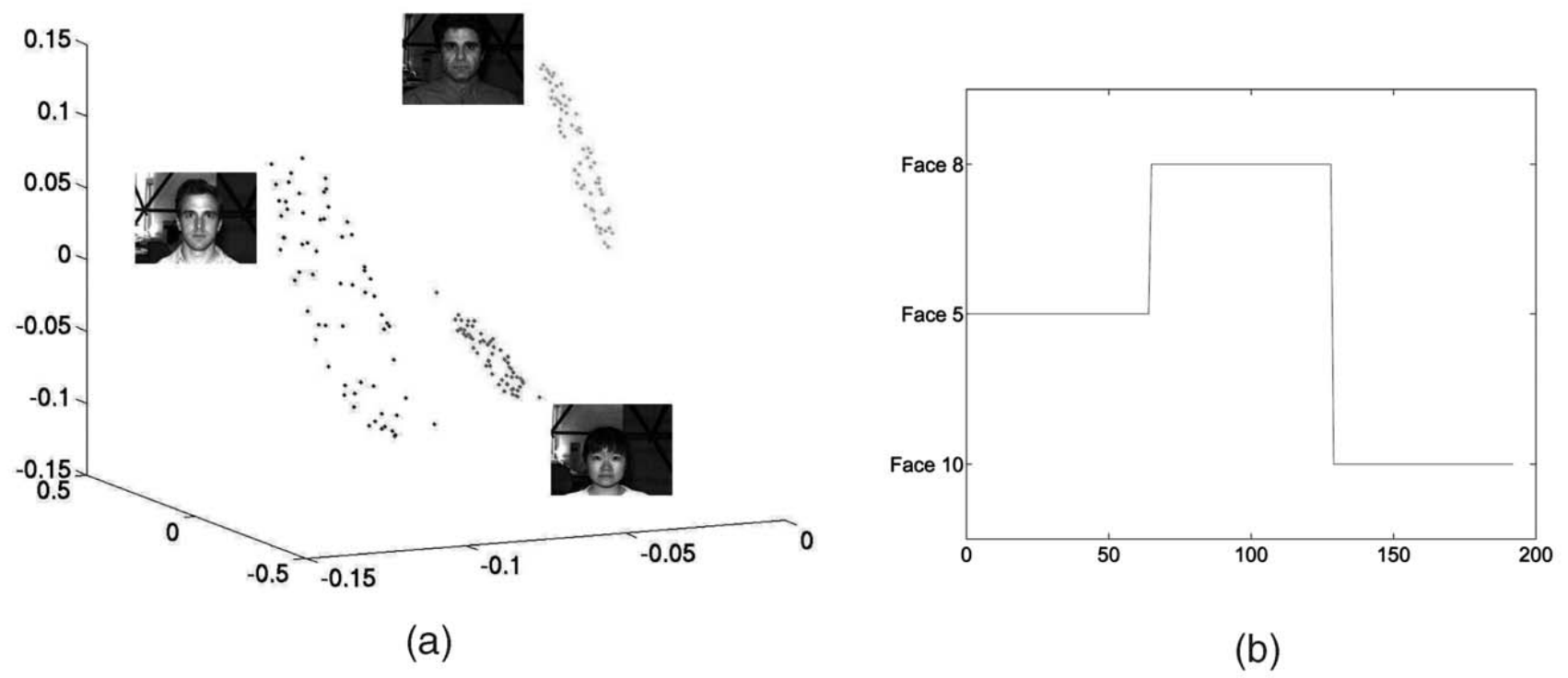

Fig. 6. Clustering a subset of the Yale Face Database B consisting of 64 frontal views under varying lighting conditions for subjects 2, 5, and 8. (a) Image data projected onto the three principal components. (b) Clustering results given by GPCA.

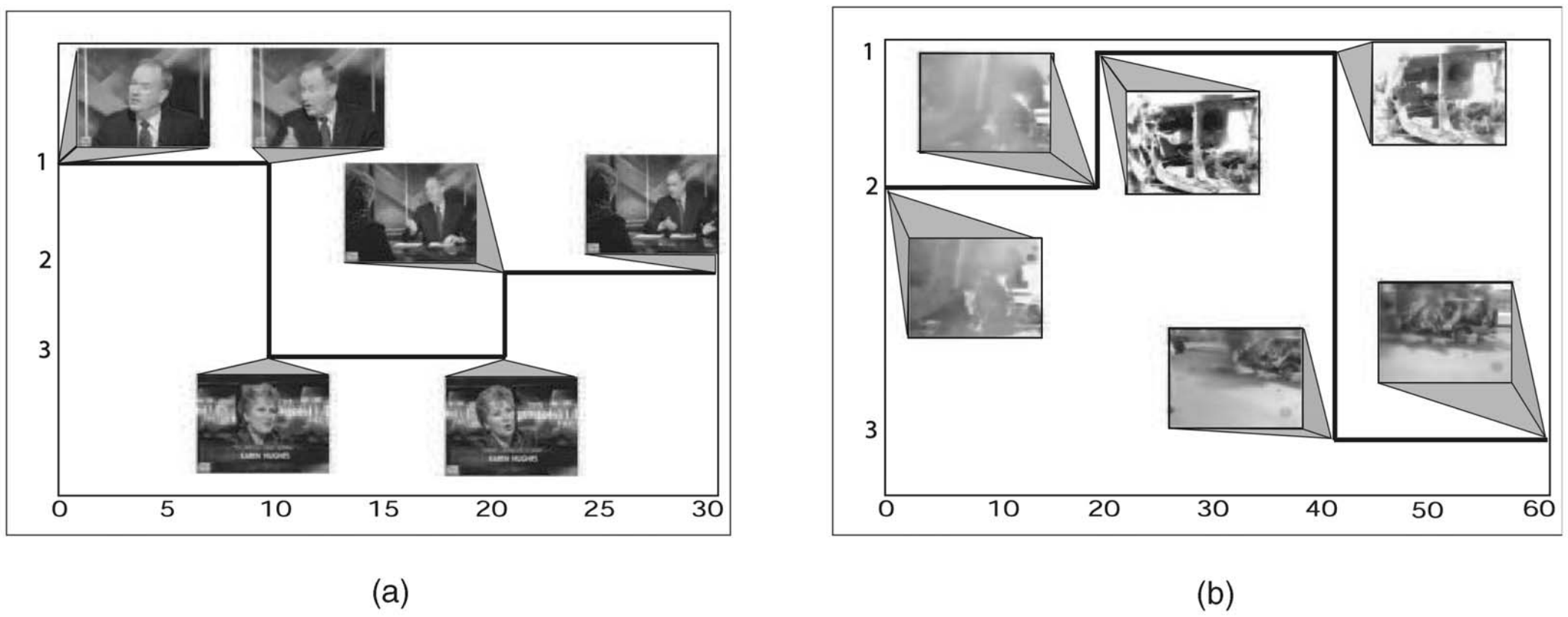

Fig. 7. Clustering frames of video sequences into groups of scenes using GPCA. (a) Thirty frames of a TV show clustered into three groups: interviewer, interviewee, and both of them. (b) Sixty frames of a sequence from Iraq clustered into three groups: rear of a car with a burning wheel, a burned car with people, and a burning car.

Table 1 shows the mean computing time and the mean number of iterations for a MATLAB implementation of each one of the algorithms over 1,000 trials. Among the algebraic algorithms, the fastest one is PFA which directly factors $p_n(\boldsymbol{x})$ given $\boldsymbol{c}_n$. The extra cost of GPCA relative to the PFA is to compute the derivatives $Dp_n(\boldsymbol{x})$ for all $\boldsymbol{x} \in \boldsymbol{X}$ and to divide the polynomials. Overall, GPCA gives about half the error of PFA in about twice as much time. Notice also that GPCA reduces the number of iterations of K-subspaces and EM to approximately 1/3 and 1/2, respectively. The computing times for K-subspaces and EM are also reduced including the extra time spent on initialization with GPCA or GPCA + K-subspaces.

### 5.2 Face Clustering under Varying Illumination

Given a collection of unlabeled images $\{I_j \in \mathbb{R}^D\}_{j=1}^N$ of $n$ different faces taken under varying illumination, we would like to cluster the images corresponding to the face of the same person. For a Lambertian object, it has been shown that the set of all images taken under all lighting conditions forms a cone in the image space, which can be well approximated by a low-dimensional subspace [10]. Therefore, we can cluster the collection of images by estimating a basis for each one of those subspaces, because images of different faces will lie in different subspaces. Since, in practice, the number of pixels $D$ is large compared with the dimension of the subspaces, we first apply PCA to project the images onto $\mathbb{R}^{D'}$ with $D' \ll D$ (see Section 4.1). More specifically, we compute the SVD of the data $[I_1\ I_2 \cdots I_N]_{D\times N} = U\Sigma V^T$ and consider a matrix $X \in \mathbb{R}^{D'\times N}$ consisting of the first $D'$ columns of $V^T$. We obtain a new set of data points in $\mathbb{R}^{D'}$ from each one of the columns of $X$. We use homogeneous coordinates $\{\boldsymbol{x}_j \in \mathbb{R}^{D'+1}\}_{j=1}^N$ so that each projected subspace goes through the origin. We consider a subset of the Yale Face Database B consisting of $N = 64n$ frontal views of $n = 3$ faces (subjects 5, 8, and 10) under $64$ varying lighting conditions. For computational efficiency, we downsampled each image to $D = 30 \times 40$ pixels. Then, we projected the data onto the first $D' = 3$ principal components, as shown in Fig. 6. We applied GPCA to the data in homogeneous coordinates and fitted three linear subspaces of dimensions 3, 2, and 2. GPCA obtained a perfect segmentation as shown in Fig. 6b.

### 5.3 Temporal Segmentation of Video Sequences

Consider a news video sequence in which the camera is switching among a small number of scenes. For instance, the host could be interviewing a guest and the camera may be switching between the host, the guest, and both of them, as shown in Fig. 7a. Given the frames $\{I_j \in \mathbb{R}^D\}_{j=1}^N$, we would like to cluster them according to the different scenes. We assume that all the frames corresponding to the same scene live in a low-dimensional subspace of $\mathbb{R}^D$ and that different scenes correspond to different subspaces. As in the case of face clustering, we may segment the video sequence into different scenes by applying GPCA to the image data projected onto the first few principal components. Fig. 7b shows the segmentation results for two video sequences. In both cases, a perfect segmentation is obtained.

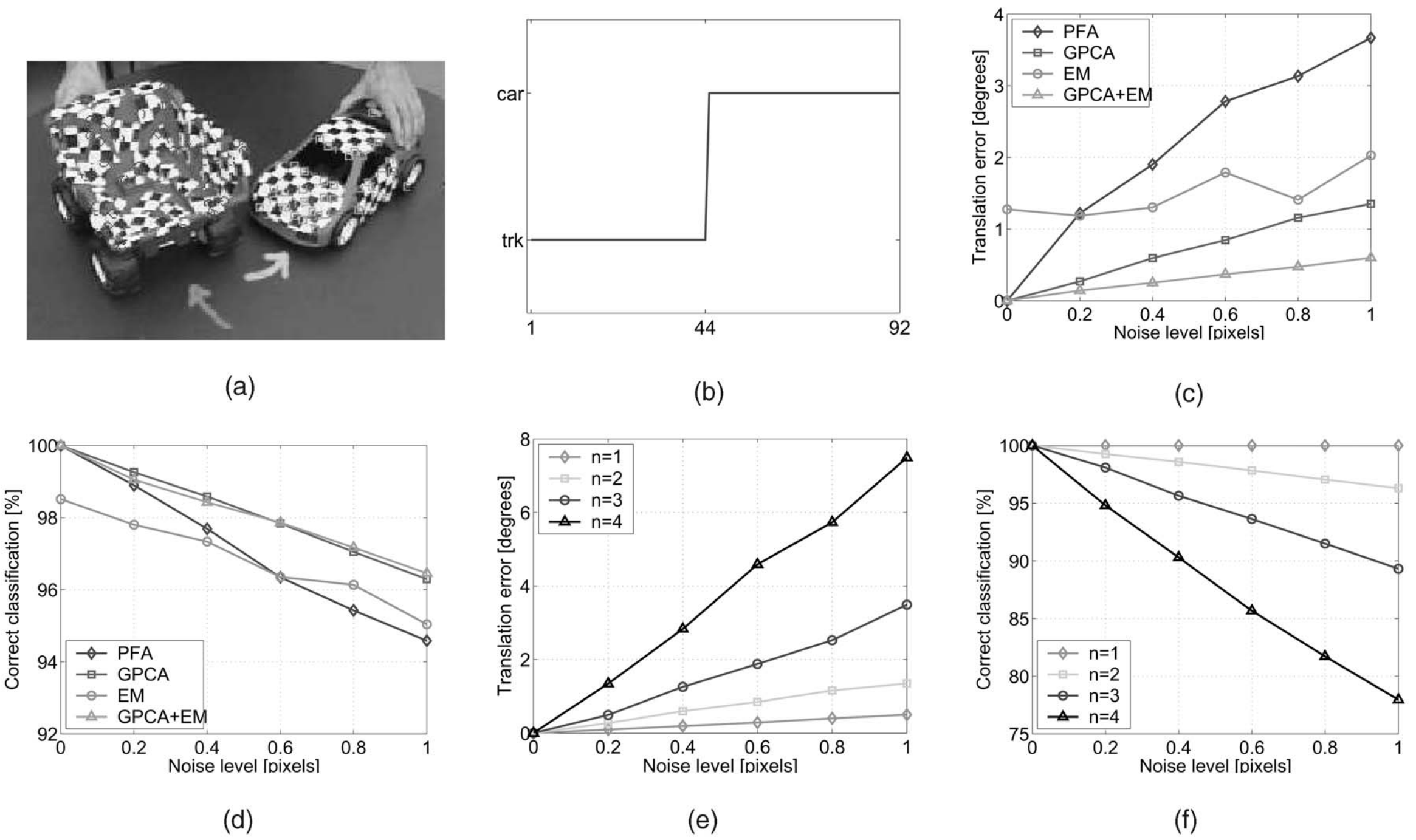

Fig. 8. Segmenting 3D translational motions by segmenting planes in $\mathbb{R}^3$. (a) First frame of a real sequence with two moving objects with 92 feature points superimposed. (b) Segmentation of the 92 feature points into two motions. (c) Error in translation and (d) percentage of correct classification of GPCA, PFA, and EM as a function of noise in the image features for $n = 2$ motions. (e) Error in translation and (f) percentage of correct classification of GPCA as a function of the number of motions.

### 5.4 Segmentation of Linearly Moving Objects

In this section, we apply GPCA to the problem of segmenting the 3D motion of multiple objects undergoing a purely translational motion. We refer the reader to [25], [26], where for the case of arbitrary rotation and translation via the segmentation of a mixture of fundamental matrices.

We assume that the scene can be modeled as a mixture of purely translational motion models, $\{T_i\}_{i=1}^n$, where $T_i \in \mathbb{R}^3$ represents the translation of object $i$ relative to the camera between the two consecutive frames. Given the images $\boldsymbol{x}_1$ and $\boldsymbol{x}_2$ of a point in object $i$ in the first and second frame, respectively, the rays $\boldsymbol{x}_1$, $\boldsymbol{x}_2$ and $T_i$ are coplanar. Therefore $\boldsymbol{x}_1$, $\boldsymbol{x}_2$ and $T_i$ must satisfy the well-known epipolar constraint for linear motions

$$\boldsymbol{x}_2^T(T_i \times \boldsymbol{x}_1) = 0. \tag{35}$$

In the case of an uncalibrated camera, the epipolar constraint reads $\boldsymbol{x}_2^T(\boldsymbol{e}_i \times \boldsymbol{x}_1) = 0$, where $\boldsymbol{e}_i \in \mathbb{R}^3$ is known as the *epipole* and is linearly related to the translation vector $T_i \in \mathbb{R}^3$. Since the epipolar constraint can be conveniently rewritten as

$$\boldsymbol{e}_i^T(\boldsymbol{x}_2 \times \boldsymbol{x}_1) = 0, \tag{36}$$

where $\boldsymbol{e}_i \in \mathbb{R}^3$ represents the epipole associated with the $i$th motion, $i = 1, \dots, n$, if we define the *epipolar line* $\boldsymbol{\ell} = (\boldsymbol{x}_2 \times \boldsymbol{x}_1) \in \mathbb{R}^3$ as a data point, then we have that $\boldsymbol{e}_i^T\boldsymbol{\ell} = 0$. Therefore, the segmentation of a set of images $\{(\boldsymbol{x}_1^j, \boldsymbol{x}_2^j)\}_{j=1}^N$ of a collection of $N$ points in 3D undergoing $n$ *distinct* linear motions $\boldsymbol{e}_1, \dots, \boldsymbol{e}_n \in \mathbb{R}^3$, can be interpreted as a subspace segmentation problem with $d = 2$ and $D = 3$, where the epipoles $\{\boldsymbol{e}_i\}_{i=1}^n$ are the normal to the planes and the epipolar lines $\{\boldsymbol{\ell}^j\}_{j=1}^N$ are the data points. One can use (26) and Algorithm 1 to determine the number of motions $n$ and the epipoles $\mathbf{e}_i$, respectively.

Fig. 8a shows the first frame of a $320 \times 240$ video sequence containing a truck and a car undergoing two 3D translational motions. We applied GPCA with $D = 3$, and $\delta = 0.02$ to the epipolar lines obtained from a total of $N = 92$ features, 44 in the truck and 48 in the car. The algorithm obtained a perfect segmentation of the features, as shown in Fig. 8b, and estimated the epipoles with an error of 5.9 degrees for the truck and 1.7 degrees for the car.

We also tested the performance of GPCA on synthetic point correspondences corrupted with zero-mean Gaussian noise with s.t.d. between 0 and 1 pixels for an image size of $500 \times 500$ pixels. For comparison purposes, we also implemented the PFA and the EM algorithm for segmenting hyperplanes in $\mathbb{R}^3$. Figs. 8c and 8d show the performance of all the algorithms as a function of the level of noise for $n = 2$ moving objects. The performance measures are the mean error between the estimated and the true epipoles (in degrees) and the mean percentage of correctly segmented feature points using 1,000 trials for each level of noise. Notice that GPCA gives an error of less than 1.3 degrees and a classification performance of over 96 percent. Thus, GPCA gives approximately 1/3 the error of PFA and improves the classification performance by about 2 percent. Notice also that EM with the normal vectors initialized at random (EM) yields a nonzero error in the noise free case, because it frequently converges to a local minimum. In fact, our algorithm outperforms EM. However, if we use GPCA to initialize EM (GPCA + EM), the performance of both algorithms improves, showing that our algorithm can be effectively used to initialize iterative approaches to motion segmentation. Furthermore, the number of iterations of GPCA + EM is approximately 50 percent with respect to EM randomly initialized; hence, there is also a gain in computing time. Figs. 8e and 8f show the performance of GPCA as a function of the number of moving objects for different levels of noise. As expected, the performance deteriorates as the number of moving objects increases, though the translation error is still below 8 degrees and the percentage of correct classification is over 78 percent.

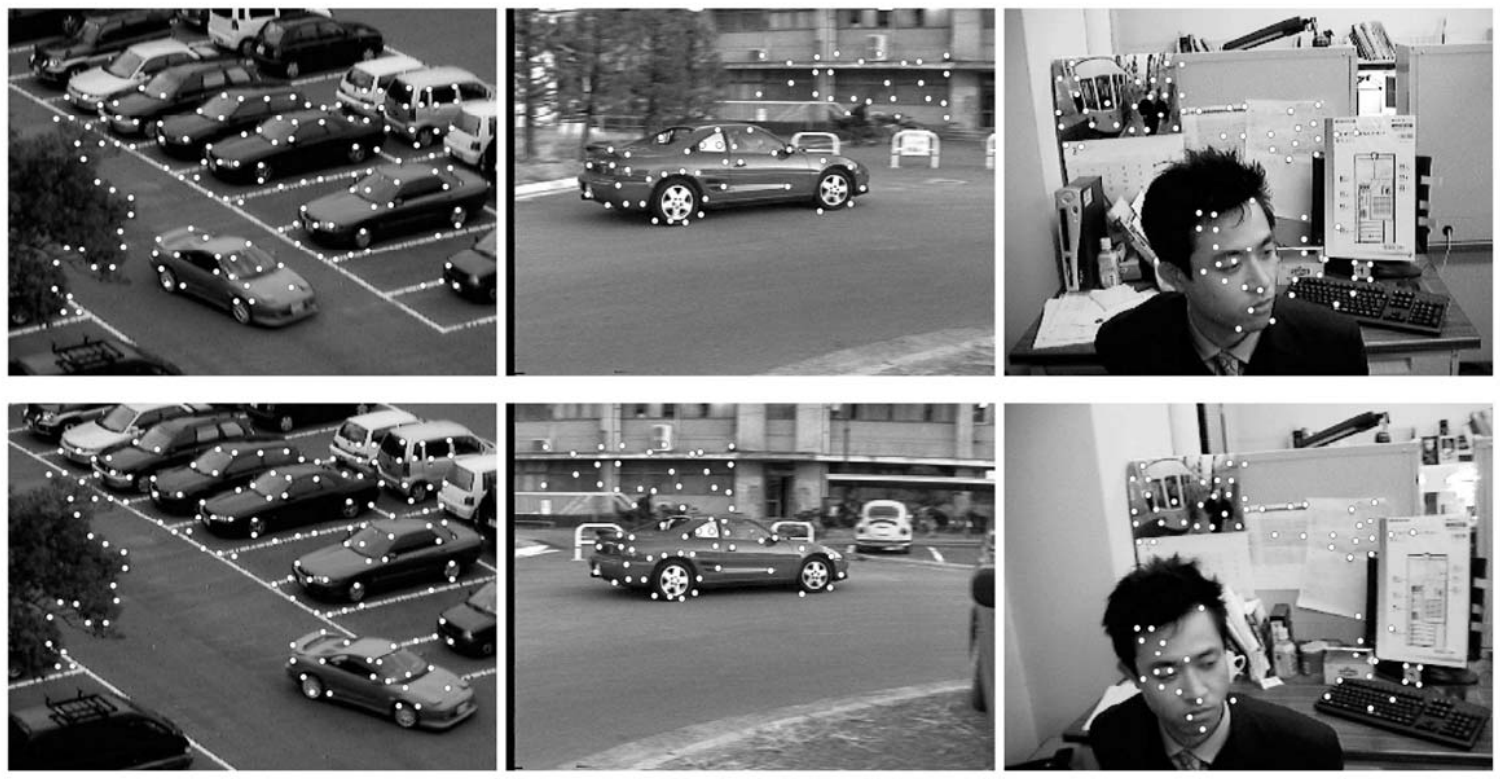

Fig. 9. Segmenting the point correspondences of sequences A (left), B (center), and C (right) in [14] for each pair of consecutive frames by segmenting subspaces in $\mathbb{R}^5$. First row: first frame of the sequence with point correspondences superimposed. Second row: last frame of the sequence with point correspondences superimposed.

### 5.5 Three-Dimensional Motion Segmentation from Multiple Affine Views

Let $\{\boldsymbol{x}_{fp} \in \mathbb{R}^2\}_{f=1,\dots,F}^{p=1,\dots,N}$ be a collection of $F$ images of $N$ 3D points $\{\boldsymbol{X}_p \in \mathbb{R}^3\}_{j=1}^N$ taken by a moving affine camera. Under the affine camera model, which generalizes orthographic, weak perspective, and paraperspective projection, the images satisfy the equation

$$\boldsymbol{x}_{fp} = A_f \boldsymbol{X}_p, \tag{37}$$

where $A_f \in \mathbb{R}^{2\times 4}$ is the affine camera matrix for frame $f$, which depends on the position and orientation of the camera as well as the internal calibration parameters. Therefore, if we stack all the image measurements into a $2F \times N$ matrix $W$, we obtain

$$W = MS^T$$
$$\begin{bmatrix} \boldsymbol{x}_{11} & \cdots & \boldsymbol{x}_{1N} \\ \vdots & & \vdots \\ \boldsymbol{x}_{F1} & \cdots & \boldsymbol{x}_{FN} \end{bmatrix}_{2F\times N} = \begin{bmatrix} A_1 \\ \vdots \\ A_F \end{bmatrix}_{2F\times 4} [\boldsymbol{X}_1 \ \cdots \ \boldsymbol{X}_N]_{4\times N}. \tag{38}$$

It follows from (38) that $\operatorname{rank}(W) \le 4$; hence, the 2D trajectories of the image points across multiple frames, that is, the columns of $W$, live in a subspace of $\mathbb{R}^{2F}$ of dimension 2, 3, or 4 spanned by the columns of the *motion matrix* $M \in \mathbb{R}^{2F\times 4}$.

Consider now the case in which the set of points $\{\boldsymbol{X}_p\}_{p=1}^N$ corresponds to $n$ moving objects undergoing $n$ different motions. In this case, each moving object spans a different $d$-dimensional subspace of $\mathbb{R}^{2F}$, where $d = 2$, 3, or 4. Solving the motion segmentation problem is hence equivalent to finding a basis for each one of such subspaces without knowing which points belong to which subspace. Therefore, we can apply GPCA to the image measurements projected onto a subspace of $\mathbb{R}^{2F}$ of dimension $D = d_{\max} + 1 = 5$. That is, if $W = U\Sigma V^T$ is the SVD of the data matrix, then we can solve the motion segmentation problem by applying GPCA to the first five columns of $V^T$.

We tested GPCA on two outdoor sequences taken by a moving camera tracking a car moving in front of a parking lot and a building (sequences A and B), and one indoor sequence taken by a moving camera tracking a person moving his head (sequence C), as shown in Fig. 9. The data for these sequences are taken from [14] and consist of point correspondences in multiple views, which are available at http://www.suri.it.okayama-u.ac.jp/data.html. For all sequences, the number of motions is correctly estimated from (11) as $n = 2$ for all values of $\kappa \in [2, 20]\,10^{-7}$. Also, GPCA gives a percentage of correct classification of 100.0 percent for all three sequences, as shown in Table 2. The table also shows results reported in [14] from existing *multiframe* algorithms for motion segmentation. The comparison is somewhat unfair, because our algorithm is purely algebraic, while the others use iterative refinement to deal with noise. Nevertheless, the only algorithm having a comparable performance to ours is Kanatani's multistage optimization algorithm, which is based on solving a series of EM-like iterative optimization problems, at the expense of a significant increase in computation.

## 6 Conclusions and Open Issues

We have proposed an algebro-geometric approach to subspace segmentation called *Generalized Principal Component Analysis* (GPCA). Our approach is based on estimating a collection of polynomials from data and then evaluating their derivatives at a data point in order to obtain a basis for the subspace passing through that point. Our experiments showed that GPCA gives about half of the error with respect to existing algebraic algorithms based on polynomial factorization, and significantly improves the performance of iterative techniques such as K-subspaces and EM. We also demonstrated the performance of GPCA on vision problems such as face clustering and video/motion segmentation.

TABLE 2
Classification Rates Given by Various Subspace Segmentation Algorithms for Sequences A, B, and C in [14]

| Sequence | A | B | C |
|---|---|---|---|
| Number of points | 136 | 63 | 73 |
| Number of frames | 30 | 17 | 100 |
| Costeira-Kanade | 60.3% | 71.3% | 58.8% |
| Ichimura | 92.6% | 80.1% | 68.3% |
| Kanatani: subspace separation | 59.3% | 99.5% | 98.9% |
| Kanatani: affine subspace separation | 81.8% | 99.7% | 67.5% |
| Kanatani: multi-stage optimization | 100.0% | 100.0% | 100.0% |
| GPCA | 100.0% | 100.0% | 100.0% |

At present, GPCA works well when the number and the dimensions of the subspaces are small, but the performance deteriorates as the number of subspaces increases. This is because GPCA starts by estimating a collection of polynomials in a linear fashion, thus neglecting the nonlinear constraints among the coefficients of those polynomials, the so-called Brill's equations [6]. Another open issue has to do with the estimation of the number of subspaces $n$ and their dimensions $\{d_i\}_{i=1}^n$ by harnessing additional algebraic properties of the vanishing ideals of subspace arrangements (e.g., the Hilbert function of the ideals). Throughout the paper, we hinted at a connection between GPCA and Kernel Methods, e.g., the Veronese map gives an embedding that satisfies the modeling assumptions of KPCA (see Remark 1). Further connections between GPCA and KPCA are worthwhile investigating. Finally, the current GPCA algorithm does not assume the existence of outliers in the given sample data, though one can potentially incorporate statistical methods such as influence theory and random sampling consensus to improve its robustness. We will investigate these problems in future research.

## Acknowledgments

The authors would like to thank Drs. Jacopo Piazzi and Kun Huang for their contributions to this work, and Drs. Frederik Schaffalitzky and Robert Fossum for insightful discussions on the topic. This work was partially supported by Hopkins WSE startup funds, UIUC ECE startup funds, and by grants NSF CAREER IIS-0347456, NSF CAREER IIS-0447739, NSF CRS-EHS-0509151, ONR YIP N00014-05-1-0633, ONR N00014-00-1-0621, ONR N000140510836, and DARPA F33615-98-C-3614.

## References

[1] D.S. Broomhead and M. Kirby, "A New Approach to Dimensionality Reduction Theory and Algorithms," *SIAM J. Applied Math.*, vol. 60, no. 6, pp. 2114-2142, 2000.

[2] M. Collins, S. Dasgupta, and R. Schapire, "A Generalization of Principal Component Analysis to the Exponential Family," *Advances on Neural Information Processing Systems*, vol. 14, 2001.

[3] J. Costeira and T. Kanade, "A Multibody Factorization Method for Independently Moving Objects," *Int'l J. Computer Vision*, vol. 29, no. 3, pp. 159-179, 1998.

[4] C. Eckart and G. Young, "The Approximation of One Matrix by Another of Lower Rank," *Psychometrika*, vol. 1, pp. 211-218, 1936.

[5] M.A. Fischler and R.C. Bolles, "RANSAC Random Sample Consensus: A Paradigm for Model Fitting with Applications to Image Analysis and Automated Cartography," *Comm. ACM*, vol. 26, pp. 381-395, 1981.

[6] I.M. Gelfand, M.M. Kapranov, and A.V. Zelevinsky, *Discriminants, Resultants, and Multidimensional Determinants.* Birkhäuser, 1994.

[7] J. Harris, *Algebraic Geometry: A First Course.* Springer-Verlag, 1992.

[8] R. Hartshorne, *Algebraic Geometry.* Springer, 1977.

[9] M. Hirsch, *Differential Topology.* Springer, 1976.

[10] J. Ho, M.-H. Yang, J. Lim, K.-C. Lee, and D. Kriegman, "Clustering Apperances of Objects under Varying Illumination Conditions," *Proc. IEEE Conf. Computer Vision and Pattern Recognition*, vol. 1, pp. 11-18, 2003.

[11] K. Huang, Y. Ma, and R. Vidal, "Minimum Effective Dimension for Mixtures of Subspaces: A Robust GPCA Algorithm and Its Applications," *Proc. IEEE Conf. Computer Vision and Pattern Recognition*, vol. 2, pp. 631-638, 2004.

[12] I. Jolliffe, *Principal Component Analysis.* New York: Springer-Verlag, 1986.

[13] K. Kanatani, "Motion Segmentation by Subspace Separation and Model Selection," *Proc. IEEE Int'l Conf. Computer Vision*, vol. 2, pp. 586-591, 2001.

[14] K. Kanatani and Y. Sugaya, "Multi-Stage Optimization for Multi-Body Motion Segmentation," *Proc. Australia-Japan Advanced Workshop Computer Vision*, pp. 335-349, 2003.

[15] A. Leonardis, H. Bischof, and J. Maver, "Multiple Eigenspaces," *Pattern Recognition*, vol. 35, no. 11, pp. 2613-2627, 2002.

[16] B. Scholkopf, A. Smola, and K.-R. Muller, "Nonlinear Component Analysis as a Kernel Eigenvalue Problem," *Neural Computation*, vol. 10, pp. 1299-1319, 1998.

[17] M. Shizawa and K. Mase, "A Unified Computational Theory for Motion Transparency and Motion Boundaries Based on Eigenenergy Analysis," *Proc. IEEE Conf. Computer Vision and Pattern Recognition*, pp. 289-295, 1991.

[18] H. Stark and J.W. Woods, *Probability and Random Processes with Applications to Signal Processing*, third ed. Prentice Hall, 2001.

[19] M. Tipping and C. Bishop, "Mixtures of Probabilistic Principal Component Analyzers," *Neural Computation*, vol. 11, no. 2, pp. 443-482, 1999.

[20] M. Tipping and C. Bishop, "Probabilistic Principal Component Analysis," *J. Royal Statistical Soc.*, vol. 61, no. 3, pp. 611-622, 1999.

[21] P. Torr, R. Szeliski, and P. Anandan, "An Integrated Bayesian Approach to Layer Extraction from Image Sequences," *IEEE Trans. Pattern Analysis and Machine Intelligence*, vol. 23, no. 3, pp. 297-303, Mar. 2001.

[22] R. Vidal and R. Hartley, "Motion Segmentation with Missing Data by PowerFactorization and Generalized PCA," *Proc. IEEE Conf. Computer Vision and Pattern Recognition*, vol. II, pp. 310-316, 2004.

[23] R. Vidal, Y. Ma, and J. Piazzi, "A New GPCA Algorithm for Clustering Subspaces by Fitting, Differentiating, and Dividing Polynomials," *Proc. IEEE Conf. Computer Vision and Pattern Recognition*, vol. I, pp. 510-517, 2004.

[24] R. Vidal, Y. Ma, and S. Sastry, "Generalized Principal Component Analysis (GPCA)," *Proc. IEEE Conf. Computer Vision and Pattern Recognition*, vol. I, pp. 621-628, 2003.

[25] R. Vidal, Y. Ma, S. Soatto, and S. Sastry, "Two-View Multibody Structure from Motion," *Int'l J. Computer Vision*, to be published in 2006.

[26] R. Vidal and Y. Ma, "A Unified Algebraic Approach to 2-D and 3-D Motion Segmentation," *Proc. European Conf. Computer Vision*, pp. 1-15, 2004.

[27] Y. Wu, Z. Zhang, T.S. Huang, and J.Y. Lin, "Multibody Grouping via Orthogonal Subspace Decomposition," *Proc. IEEE Conf. Computer Vision and Pattern Recognition*, vol. 2, pp. 252-257, 2001.

[28] L. Zelnik-Manor and M. Irani, "Degeneracies, Dependencies and Their Implications in Multi-Body and Multi-Sequence Factorization," *Proc. IEEE Conf. Computer Vision and Pattern Recognition*, vol. 2, pp. 287-293, 2003.

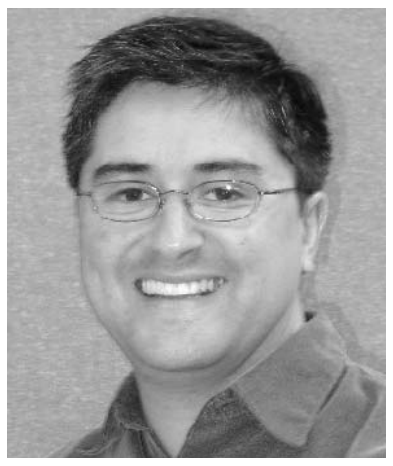

**René Vidal** received the BS degree in electrical engineering (highest honors) from the Universidad Católica de Chile in 1997, and the MS and PhD degrees in electrical engineering and computer sciences from the University of California at Berkeley in 2000 and 2003, respectively. In 2004, he joined The Johns Hopkins University as an assistant professor in the Department of Biomedical Engineering and the Center for Imaging Science. His areas of research are biomedical imaging, computer vision, machine learning, hybrid systems, robotics, and vision-based control. Dr. Vidal is recipient of the 2005 CAREER Award from the US National Science Foundation, the 2004 Best Paper Award Honorable Mention at the European Conference on Computer Vision, the 2004 Sakrison Memorial Prize, the 2003 Eli Jury Award, and the 1997 Award of the School of Engineering of the Universidad Católica de Chile to the best graduating student of the school. He is a member of the IEEE.

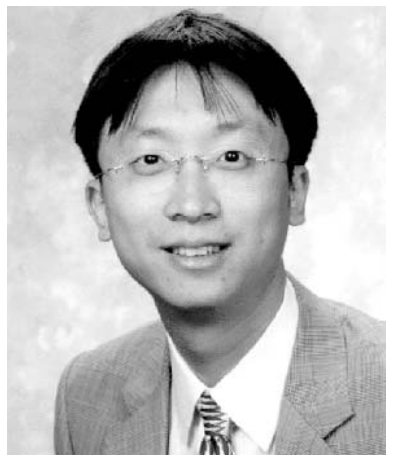

**Yi Ma** received two bachelors' degrees in automation and applied mathematics from Tsinghua University, Beijing, China, in 1995. He received the MS degree in electrical engineering and computer science (EECS) in 1997, the MA degree in mathematics in 2000, and the PhD degree in EECS in 2000 all from the University of California. Since 2000, he has been an assistant professor in the Electrical and Computer Engineering Department at the University of Illinois at Urbana-Champaign (UIUC). He has coauthored more than 40 technical papers and is the first author of a book, entitled *An Invitation to 3-D Vision: From Images to Geometric Models*, published by Springer, 2003. Dr. Ma was the recipient of the David Marr Best Paper Prize at the International Conference on Computer Vision in 1999 and honorable mention for the Longuet-Higgins Best Paper Award at the European Conference on Computer Vision in 2004. He received the CAREER Award from the US National Science Foundation in 2004 and the Young Investigator Program Award from the US Office of Naval Research in 2005. He is a member of the IEEE.

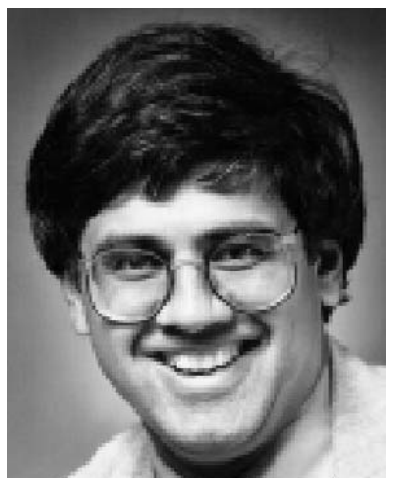

**Shankar Sastry** received the MA degree (honoris causa) from Harvard in 1994 and the PhD degree in 1981 from the University of California, Berkeley. He was on the faculty of Massachusetts Institute of Technology (MIT) as an assistant professor from 1980 to 1982 and Harvard University as a chaired Gordon Mc Kay professor in 1994. He served as chairman of the Electrical Engineering and Computer Science (EECS) Department, University of California, Berkeley from 2001 to 2004. In 2000, he served as director of the Information Technology Office at DARPA. He is the NEC Distinguished Professor of EECS and a professor of bioengineering and currently serves as the director of Center for Information Technology in the Interests of Society (CITRIS). He has coauthored more than 300 technical papers and nine books. He received the president of India Gold Medal in 1977, the IBM Faculty Development award for 1983-1985, the US National Science Foundation Presidential Young Investigator Award in 1985, the Eckman Award of the American Automatic Control Council in 1990, the distinguished Alumnus Award of the Indian Institute of Technology in 1999, the David Marr prize for the best paper at the International Conference in Computer Vision in 1999, and the Ragazzini Award for Excellence in Education by the American Control Council in 2005. He is a member of the National Academy of Engineering and the American Academy of Arts and Sciences, and he became a fellow of the IEEE in 1994. He is on the US Air Force Science Board and is chairman of the Board of the International Computer Science Institute. He is also a member of the boards of the Federation of American Scientists and ESCHER (Embedded Systems Consortium for Hybrid and Embedded Research).